\documentclass[a4paper,fleqn]{cas-dc}

\usepackage[numbers,sort&compress]{natbib}
\usepackage{graphicx}
\usepackage{amsthm,amsmath,amsfonts,amssymb}
\usepackage{mathrsfs}
\usepackage{booktabs}
\usepackage{algorithm} %format of the algorithm 
\usepackage{algorithmic} %format of the algorithm 
\usepackage{subfig}
\usepackage{multirow}
\usepackage{multicol}
\usepackage{xcolor}
\usepackage{diagbox}
\usepackage{bbding}
\usepackage{makecell}
\usepackage{picins}
\definecolor{ccr}{RGB}{10,110,150}
\usepackage{hyperref}
\hypersetup{hypertex=true,colorlinks=true,linkcolor=ccr,anchorcolor=ccr,citecolor=ccr}
\usepackage{makecell}
\usepackage{manyfoot}
\usepackage{stfloats}

\def\tsc#1{\csdef{#1}{\textsc{\lowercase{#1}}\xspace}}
\tsc{WGM}
\tsc{QE}
\begin{document}
\let\WriteBookmarks\relax
\def\floatpagepagefraction{1}
\def\textpagefraction{.001}

% Short title

% Short title
\shorttitle{FPDIoU Loss: A Loss Function for Efficient Bounding Box Regression of Rotated Object Detection}    

% Short author
\shortauthors{S. Ma, Y.Xu}   

% Main title of te paper
\title [mode = title]{FPDIoU Loss: A Loss Function for Efficient Bounding Box Regression of Rotated Object Detection}  

% Title footnote mark
% eg: \tnotemark[1]
% \tnotemark[<tnote number>] 

% Title footnote 1.
% eg: \tnotetext[1]{Title footnote text}
% \tnotetext[<tnote number>]{<tnote text>} 

% First author
%
% Options: Use if required
% eg: \author[1,3]{Author Name}[type=editor,
%       style=chinese,
%       auid=000,
%       bioid=1,
%       prefix=Sir,
%       orcid=0000-0000-0000-0000,
%       facebook=<facebook id>,
%       twitter=<twitter id>,
%       linkedin=<linkedin id>,
%       gplus=<gplus id>]
\author[1]{Siliang Ma}
[orcid=0000-0001-7117-9333]

% Corresponding author indication

% Footnote of the first author
% \fnmark[The corresponding author]

% Email id of the first author
\ead{1}

% URL of the first author
\ead[url]{202010107394@mail.scut.edu.cn}

% Credit authorship
% eg: \credit{Conceptualization of this study, Methodology, Software}
\credit{Conceptualization, Methodology, Programming, Data creation, Writing - original draft}

% Address/affiliation
\affiliation[1]{organization={Institute of Computer Science and Engineering, South China University of Technology},
            addressline={No. 382, Outer Ring East Road, Panyu District}, 
            city={Guangzhou},
%          citysep={}, % Uncomment if no comma needed between city and postcode
            postcode={510006}, 
            state={Guangdong},
            country={China}}

\author[1,2]{Yong Xu}
\cormark[1]
% Footnote of the second author
% \fnmark[2]

% Email id of the second author
\ead{2}

% URL of the second author
\ead[url]{yxu@scut.edu.cn}

% Credit authorship
\credit{Supervision, Writing-Reviewing and Editing}

% Address/affiliation
\affiliation[2]{organization={Pengcheng Laboratory},
            addressline={No. 2, Xingke 1st Street, Nanshan District}, 
            city={Shenzhen},
%          citysep={}, % Uncomment if no comma needed between city and postcode
            postcode={518000}, 
            state={Guangdong},
            country={China}}

% Corresponding author text
\cortext[1]{Corresponding author}

% Footnote text
% \fntext[1]{}

% For a title note without a number/mark
%\nonumnote{}

% Here goes the abstract
\begin{abstract}
Bounding box regression is one of the important steps of object detection. However, rotation detectors often involve a more complicated loss based on SkewIoU which is unfriendly to gradient-based training. Most of the existing loss functions for rotated object detection calculate the difference between two bounding boxes only focus on the deviation of area or each points distance (e.g., $\mathcal{L}_{Smooth-\ell 1}$, $\mathcal{L}_{RotatedIoU}$ and $\mathcal{L}_{PIoU}$). The calculation process of some loss functions is extremely complex (e.g. $\mathcal{L}_{KFIoU}$). In order to improve the efficiency and accuracy of bounding box regression for rotated object detection, we proposed a novel metric for arbitrary shapes comparison based on minimum points distance, which takes most of the factors from existing loss functions for rotated object detection into account, i.e., the overlap or nonoverlapping area, the central points distance and the rotation angle. We also proposed a loss function called $\mathcal{L}_{FPDIoU}$ based on four points distance for accurate bounding box regression focusing on faster and high quality anchor boxes. In the experiments, $FPDIoU$ loss has been applied to state-of-the-art rotated object detection (e.g., RTMDET, H2RBox) models training with three popular benchmarks of rotated object detection including DOTA, DIOR, HRSC2016 and two benchmarks of arbitrary orientation scene text detection including ICDAR 2017 RRC-MLT and ICDAR 2019 RRC-MLT, which achieves better performance than existing loss functions.
\end{abstract}

% Use if graphical abstract is present
%\begin{graphicalabstract}
%\includegraphics{}
%\end{graphicalabstract}

% Research highlights
\begin{highlights}
  \item Bounding box regression is a significant process of rotated object detection.
  \item Most of the existing loss functions for rotated object detection are extremely complex or with low efficiency.
  \item Geometrical properties are not fully considered to design the losses of bounding box regression.
  \item We proposed $\mathcal{L}_{FPDIoU}$ based on four points distance to tackle the issues of existing losses and with higher efficiency and accuracy.
\end{highlights}

% Keywords
% Each keyword is seperated by \sep

\maketitle

% Numbered list
% Use the style of numbering in square brackets.
% If nothing is used, default style will be taken.
%\begin{enumerate}[a)]
%\item 
%\item 
%\item 
%\end{enumerate}  

% Unnumbered list
%\begin{itemize}
%\item 
%\item 
%\item 
%\end{itemize}  

% Description list
%\begin{description}
%\item[]
%\item[] 
%\item[] 
%\end{description}  

% Figure

\section{Introduction}
Rotated object detection can be considered as a special circumstance of object detection, which has attracted a large scale of researchers' interests during the past few years.
Most of the state-of-the-art rotated object detectors can be modified based on horizontal object detectors (e.g., YOLO series \cite{2017YOLO9000,2018YOLOv3,2020YOLOv4,2022YOLOv7,YOLOF,YOLOv5}, Faster R-CNN \cite{2016Faster}, reppoints \cite{yang2019reppoints} and retinanet \cite{lin2017focal}), which rely on a bounding box regression (BBR) module to localize objects.

\begin{figure}
  \centering
  \subfloat[$\mathcal{L}_{RotatedIoU}=0.4898$,\\ $\mathcal{L}_{GIoU}=0.4898$,\\ $\mathcal{L}_{DIoU}=0.4898$,\\
  $\mathcal{L}_{PIoU}=0.4898$,\\ $\mathcal{L}_{KFIoU}=3.6563$,\\ \textcolor{red}{$\mathcal{L}_{FPDIoU}=0.5098$}]{
    \includegraphics[width=0.45\linewidth]{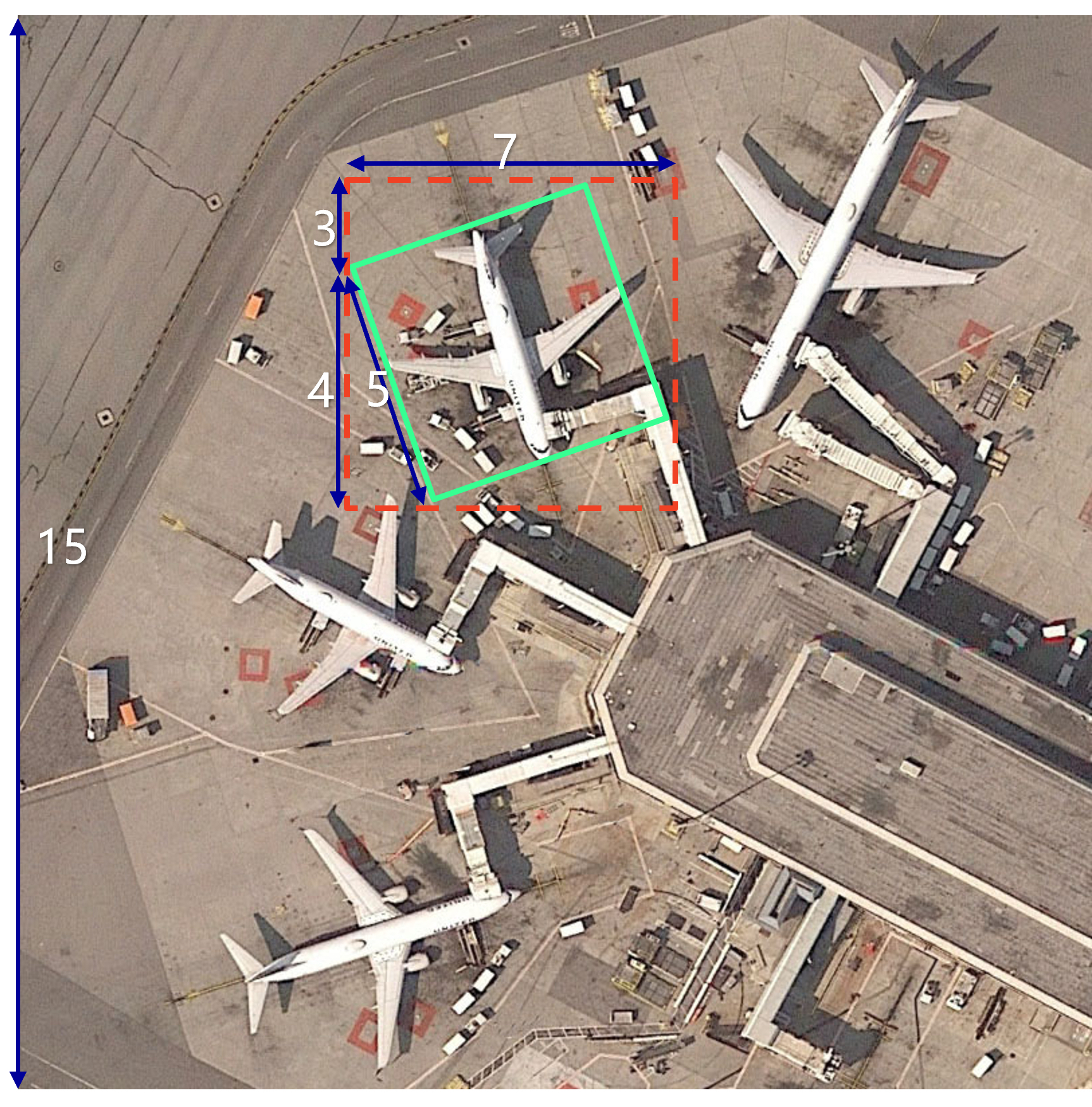}
  }
  \hfill
  \subfloat[$\mathcal{L}_{RotatedIoU}=0.4898$,\\ $\mathcal{L}_{GIoU}=0.4898$,\\ $\mathcal{L}_{DIoU}=0.4898$,\\
  $\mathcal{L}_{PIoU}=0.4898$,\\ $\mathcal{L}_{KFIoU}=3.6563$,\\
  \textcolor{red}{$\mathcal{L}_{FPDIoU}=0.4942$}]{
    \includegraphics[width=0.45\linewidth]{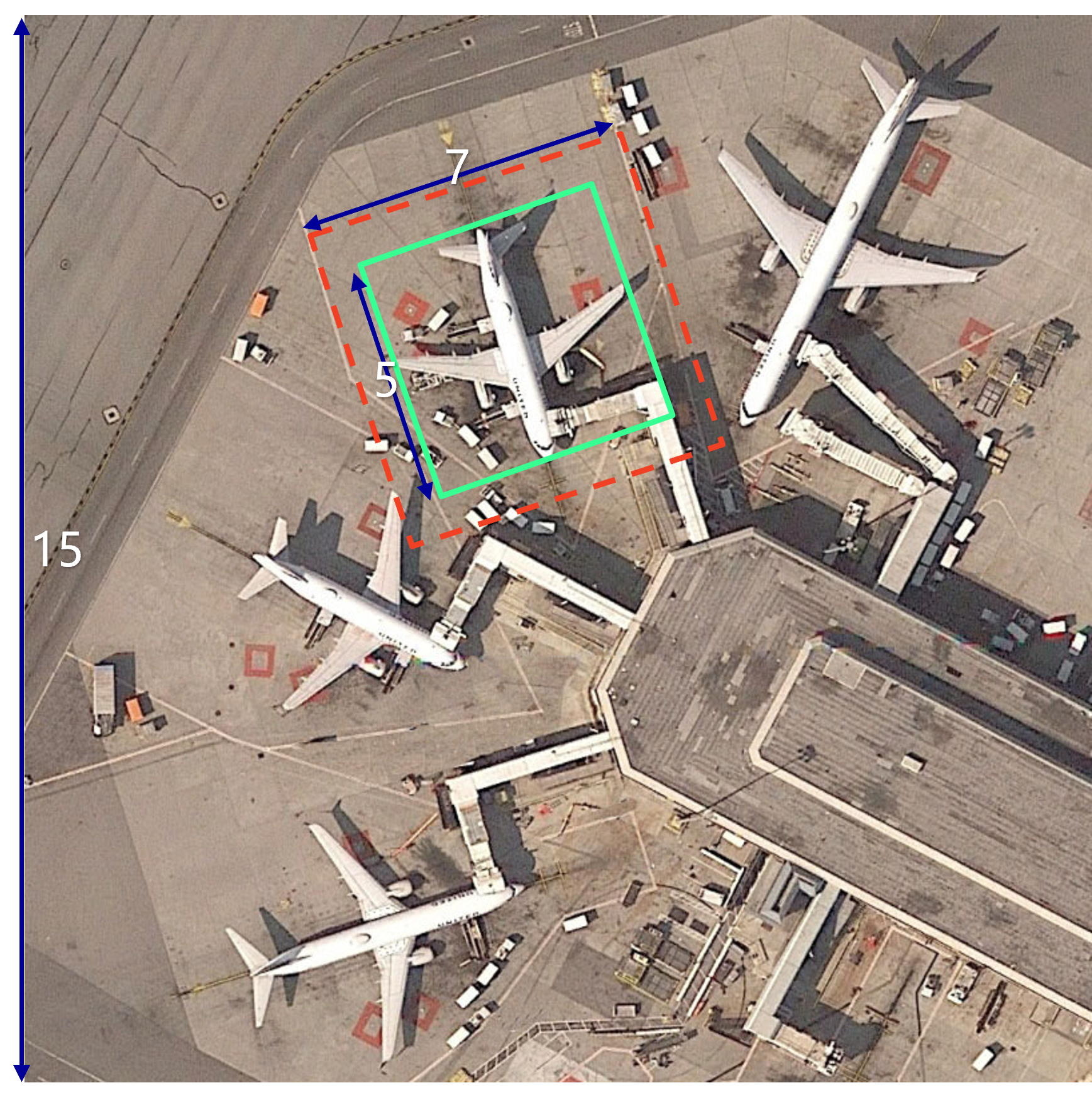}
  }
  \caption{Two cases with different bounding boxes regression results. The \textcolor{green}{green} boxes denote the groundtruth bounding boxes and the \textcolor{red}{red} boxes denote the predicted bounding boxes. The $\mathcal{L}_{RotatedIoU}$, $\mathcal{L}_{GIoU}$, $\mathcal{L}_{DIoU}$, $\mathcal{L}_{PIoU}$, $\mathcal{L}_{KFIoU}$ between these two cases are exactly same value, but their $\mathcal{L}_{FPDIoU}$ values are very different.}
  \label{intro}
\end{figure}

During the past few years, $IoU$ \cite{IoU} has been widely used for comparing the similarity between two arbitrary shapes based on the overlap area. This property makes $IoU$ invariant to the scale of the problem under consideration. Due to this appealing property, most of the performance measures used to evaluate for instance segmentation \cite{Cityscapes,2015The,ARMCV,zhou2017scene}, object detection \cite{2014Microsoft,2015The}, and tracking \cite{leal2015motchallenge} rely on this metric.

However, as Figure \ref{intro} shows, most of the existing loss functions for bounding box regression have the same value in some special cases, which will limit the convergence speed and accuracy. Therefore, we try to design a novel loss function called $\mathcal{L}_{FPDIoU}$ for bounding box regression with consideration of the advantages included in  $\mathcal{L}_{RotatedIoU}$, $\mathcal{L}_{GIoU}$, $\mathcal{L}_{DIoU}$, $\mathcal{L}_{PIoU}$ and $\mathcal{L}_{KFIoU}$, but also has higher efficiency and accuracy for bounding box regression.

In this paper, we will address the issues of the existing $IoU$-based metrics by considering the mathematical definition of rotated rectangle. We ensure this novel metric for bounding box regression follows the same definition as $IoU$, i.e. encoding the shape properties of the compared objects into the region property. In addition, it is valid whether the predicted bounding box and the groundtruth bounding box are overlap or not. We introduce this
$IoU$ based on four points distance, named $FPDIoU$, as a new metric for comparing any two arbitrary shapes. We also provide an analytical solution for calculating $FPDIoU$ between two rotated rectangles, allowing it to be used as a loss in this case. Incorporating $FPDIoU$ loss into state-of-the-art rotated object detection algorithms, we consistently improve their performance on popular object detection benchmarks such
as DOTA \cite{DOTA}, DIOR \cite{DIOR} and HRSC2016 \cite{hrsc} and arbitrary orientation scene text spotting benchmarks ICDAR 2017 RRC-MLT \cite{nayef2017icdar2017} and ICDAR 2019 RRC-MLT \cite{nayef2019icdar2019} using the
standard evaluation metric, i.e. $IoU$ based \cite{2015The,2014Microsoft} performance measure.

The contributions of this paper can be summarized as below:\\
1. We considered the advantages and disadvantages of the existing $IoU$-based losses and $\ell_{n}$-norm losses, we propose an $IoU$ loss based on four points distance called $\mathcal{L}_{FPDIoU}$ to tackle the issues of existing losses and obtain a faster convergence speed and more accurate regression results.\\
2. Extensive experiments have been conducted on rotated object detection tasks. We compared with the existing loss functions for rotated bounding box regression and the SOTA rotated object detectors.  Outstanding experimental results validate the superiority of the proposed $FPDIoU$ loss.
\section{Related Work}
\subsection{Rotated Object Detection}
Different from horizontal object detection, rotated object detection needs to consider not only the central points, width and height of the bounding boxes, but also the rotation angle. At the very beginning, researchers attempted to extend classical horizontal detectors \cite{FastR-CNN,2016Faster,2017Feature} to the rotation case by adopting the rotated bounding boxes. As we all know, aerial images and arbitrary orientation scene text spotting are popular application of rotated object detection. For aerial images, objects are always not with large aspect ratios. To this end, various rotated object detection models have been proposed. As for single stage methods, DRN \cite{DRN}, DAL \cite{ming2021dynamic}, R$^{3}$Det \cite{yang2021r3det}, RSDet and S$^{2}$A-Net \cite{s2anet} are widely used to deal with rotated object detection tasks. While ICN \cite{ICN} , ROI-Transformer \cite{RoI-trans}, SCRDet \cite{SCRDet}, Mask OBB \cite{Wang2019MaskOA}, Gliding Vertex \cite{GlidingVertex}, ReDet \cite{ReDet} are two-stage mainstreamed approaches whose pipeline is inherited from Faster RCNN \cite{2016Faster}. In order to provide more efficient and accurate rotated object detection results, RTMDET \cite{lyu2022rtmdet} was proposed to deal with real-time rotated object detection tasks, which uses dynamic soft labels to optimize training strategies. However, in order to simplify the rotated object detection problem, a weakly supervised method named H2RBox \cite{yang2022h2rbox} was proposed which only use horizontal annotations to deal with rotated object detection task. No matter we choose which rotated object detectors, the bounding box regression is necesarry, which performs significant role in rotated object detection.

\subsection{Variants of IoU-based Loss}
It is well-known that IoU-based losses have been widely used in horizontal object detection. At the very beginning, $\mathcal{L}_{n}$-norm loss function was widely used for bounding box regression, which was exactly simple but sensitive to various scales. In YOLO v1 \cite{YOLOv1}, square roots for $w$ and $h$ are adopted to mitigate this effect, while YOLO v3 \cite{2018YOLOv3} uses $2-wh$. In order to better calculate the diverse between the groundtruth and the predicted bounding boxes, $IoU$ loss is used since Unitbox \cite{2016UnitBox}. To ensure the training stability, Bounded-$IoU$ loss \cite{tychsen2018improving} introduces the upper bound of $IoU$. The original $IoU$ can be formulated as
\begin{equation}
  RotatedIoU=\frac{\mathcal{B}_{gt}\bigcap \mathcal{B}_{prd}}{\mathcal{B}_{gt}\bigcup  \mathcal{B}_{prd}},
  \label{LIoU}
\end{equation}
where $\mathcal{B}_{gt}$ denotes the groundtruth bounding box, $\mathcal{B}_{prd}$ denotes the predicted bounding box. As we can see, the $RotatedIoU$ only calculates the union area of two bounding boxes, which can't distinguish the cases that two boxes do not overlap. If the predicted bounding box completely covers the groundtruth bounding box, the $IoU$ value will not change regardless of the rotation angle.

As equation \ref{LIoU} shows, if $|\mathcal{B}_{gt}\bigcap \mathcal{B}_{prd}|=0$, then\\ $IoU(\mathcal{B}_{gt},\mathcal{B}_{prd})=0$. In this case, $IoU$ can not reflect if two shapes are in vicinity of each other or very far from each other. Then, $GIoU$ \cite{2019Generalized} is proposed to tackle this issue. The $GIoU$ can be formulated as
\begin{equation}
  GIoU=RotatedIoU-\frac{\mid \mathcal{C} -\mathcal{B}_{gt} \cup \mathcal{B}_{prd}\mid}{\mid \mathcal{C} \mid},
\end{equation}
where $\mathcal{C}$ is the smallest box covering $\mathcal{B}_{gt}$ and $\mathcal{B}_{prd}$, and $\mid \mathcal{C}\mid$ is the area of box $\mathcal{C}$. Due to the introduction of the penalty term in  $GIoU$ loss, the predicted box will move toward the target box in nonoverlapping cases.  $GIoU$ loss has been applied to train state-of-the-art object detectors, such as YOLO v3 and Faster R-CNN, and achieves better performance than MSE loss and $IoU$ loss. However,  $GIoU$ will lost effectiveness when the predicted bounding box is absolutely covered by the groundtruth bounding box. In order to deal with this problem, $DIoU$ \cite{zheng2020distance} was proposed with consideration of the centroid points distance between the predicted bounding box and the groundtruth bounding box. The formulation of $DIoU$ can be formulated as 
\begin{equation}
  DIoU=RotatedIoU-\frac{\rho ^2 (\mathcal{B}_{gt},\mathcal{B}_{prd})}{\mathcal{C} ^2},
\end{equation}
where $\rho$ denotes Euclidean distance. As we can see, the target of $\mathcal{L}_{DIoU}$ directly minimizes the distance between central points of predicted bounding box and groundtruth bounding box. However, when the central point of predicted bounding box coincides with the central point of groundtruth bounding box, it degrades to the original $IoU$. To address this issue, $CIoU$ was proposed with consideration of both central points distance and the aspect ratio. The formulation of $CIoU$ can be written as follows:
\begin{equation}
 CIoU=RotatedIoU-\frac{\rho ^2 (\mathcal{B}_{gt},\mathcal{B}_{prd})}{\mathcal{C} ^2}-\alpha V,
\end{equation}
\begin{equation}
  V =\frac{4}{\pi ^2}(\arctan \frac{w^{gt}}{h^{gt}}-\arctan \frac{w^{prd}}{h^{prd}})^2,
\end{equation}
\begin{eqnarray}
  \alpha =\frac{V}{1-IoU+V}.
\end{eqnarray}

However, the definition of aspect ratio from $CIoU$ is relative value rather than absolute value, which is not that accurate for bounding box regression. To address this issue, $EIoU$ \cite{ZHANG2022146} was proposed based on $DIoU$, which is defined as follows:

\begin{equation}
  EIoU=DIoU-\frac{\rho ^2 (w_{prd},w_{gt})}{(w^{c}) ^2}-\frac{\rho ^2 (h_{prd},h_{gt})}{(h^{c}) ^2}.
\end{equation}

According to our survey, $CIoU$ and $EIoU$ are not widely used in rotated object detection because the distance between two not parallel lines is difficult to calculate. In order to calculate the $IoU$ more accurately, $PIoU$ \cite{PIoU} was proposed based on pixel-level area calculation. The keypoint of $PIoU$ is how to judge a pixel is inside or outside the predicted bounding box and groundtruth bounding box. The formula is shown as below:

\begin{equation}
  \delta (p_{i,j}|b)=
  \begin{cases}
    1, d_{i,j}^{w}\leq \frac{w}{2}, d_{i,j}^{h}\leq \frac{h}{2}\\
    0, otherwise.\\
  \end{cases}
\end{equation}
where $d_{i,j}$ denotes the Euclidean distance between pixel $(i, j)$ and central point $(c_{x}, c_{y})$ of predicted bounding box or groundtruth bounding box, $d_{w}$ and $d_{h}$ denote the distance $d$ along horizontal and vertical direction respectively:
\begin{equation}
  d_{i,j}=\sqrt{(c_{x}-i)^2+(c_{y}-j)^2},
\end{equation}
\begin{equation}
  d_{i,j}^w=|d_{i,j}\cos \beta|,
\end{equation}
\begin{equation}
  d_{i,j}^h=|d_{i,j}\sin \beta|,
\end{equation}
\begin{equation}
  \beta =
  \begin{cases}
    \theta+\arccos \frac{c_{x}-i}{d_{ij}}, c_{y}-j\geq 0\\
    \theta-\arccos \frac{c_{x}-i}{d_{ij}}, c_{y}-j \textless 0\\
  \end{cases}.
\end{equation}
To ensure the $PIoU$ as a continuous and differentiable function, the definition of $PIoU$ is shown as below:
\begin{equation}
  K(d,s)=1-\frac{1}{1+e^{-k(d-s)}},
\end{equation}
\begin{equation}
  F(p_{i,j}|b)=K(d_{i,j}^w,w)*K(d_{i,j}^h,h),
\end{equation}

where $k$ is an adjustable factor to control the sensitivity of the target pixel $p_{i,j}$. To reduce the calculational complexity The intersection area $S_{\mathcal{B}_{gt}\bigcap \mathcal{B}_{prd}}$ and union area$S_{\mathcal{B}_{gt}\bigcup \mathcal{B}_{prd}}$ between $\mathcal{B}_{gt}$ and $\mathcal{B}_{prd}$ can be approximately defined as:

  \begin{equation}
    S_{\mathcal{B}_{gt}\bigcap \mathcal{B}_{prd}}\approx \sum \limits_{p_{i,j}\in \mathcal{B}_{gt},\mathcal{B}_{prd}}F(p_{i,j}|\mathcal{B}_{gt})F(p_{i,j}|\mathcal{B}_{prd}) ,
  \end{equation}
\begin{footnotesize}
  \begin{equation}
    S_{\mathcal{B}_{gt}\bigcup \mathcal{B}_{prd}}\approx \sum \limits_{p_{i,j}\in \mathcal{B}_{gt},\mathcal{B}_{prd}}F(p_{i,j}|\mathcal{B}_{gt})+F(p_{i,j}|\mathcal{B}_{prd})- F(p_{i,j}|\mathcal{B}_{gt})F(p_{i,j}|\mathcal{B}_{prd}),
  \end{equation}
\end{footnotesize}
\begin{equation}
    PIoU=\frac{S_{\mathcal{B}_{gt}\bigcap \mathcal{B}_{prd}}}{S_{\mathcal{B}_{gt}\bigcup \mathcal{B}_{prd}}}.
\end{equation}

On the other hand, some researchers start to think about how to make full use of Gaussian modeling. GWD\cite{Yang2021RethinkingRO} was proposed based on Gaussian Wasserstein Distance which leads to an approximate and differentiable IoU induced loss. The calculation process of GWD is formulated as:
% \begin{equation}
\begin{align}
  d^2 &=\|m_1-m_2\|_2^2+\|\Sigma _1^\frac{1}{2}-\Sigma _2^\frac{1}{2}\|_F^2 \notag\\&=(x_1-x_2)^2+(y_1-y_2)^2+\frac{(w_1-w_2)^2+(h_1-h_2)^2}{4} \notag\\&=
  \ell_2-norm([x_1,y_1,\frac{w_1}{2},\frac{h_1}{2}]^T,[x_2,y_2,\frac{w_2}{2},\frac{h_2}{2}]^T)
\end{align}
\begin{equation}
  GWD=\frac{1}{\tau +f(d^2)}, \tau \geq 1
\end{equation}
% \end{equation}

where $f(\cdot )$ denotes a non-linear function to transform the Wasserstein distance $d^2$
to make the loss more smooth and expressive. The hyperparameter $\tau$ modulates the GWD. Although GWD scheme has played a preliminary exploration of the deductive paradigm, it does not
focus on achieving high-precision detection and scale invariance. Similar with GWD, KLD \cite{KLD} was proposed with a better central point optimization mechanism, which also transforms the rotated rectangle to the 2-D Gaussian distribution. The KLD between two 2-D Gaussian can be formulated as:
\begin{equation}
  D_{kl}(\mathcal{N}_p\Vert\mathcal{N}_t)=\frac{1}{2}(\mu_p-\mu_t)^\top \Sigma_t^{-1}(\mu_p-\mu_t)+\frac{1}{2}Tr(\Sigma_t^{-1}\Sigma_p)+\frac{1}{2}ln\frac{|\Sigma_t| }{|\Sigma_p|}-1
\end{equation}
$D_{kl}(\mathcal{N}_p\Vert\mathcal{N}_t)$ is composed of partial parameter coupling, which makes all parameters form a chain coupling relationship. In the optimization process of the KLD-based detector, the parameters can influence each other and are jointly optimized, which leads to the  optimization mechanism of the model self-modulated. To make full use of the advantages of Gaussian modeling, KFIoU \cite{yang2022kfiou} was proposed based on Gaussian distribution. The volume of a specific rotated bounding box can be formulated as:
\begin{equation}
  \nu_{B}(\Sigma )=2^n\sqrt{\Pi eig(\Sigma)}=2^n|\Sigma^\frac{1}{2}|=2^n|\Sigma|^\frac{1}{2},
\end{equation}
where $n$ denotes the number of dimensions. The distribution function of the overlapping area of two Gaussian distributions $\mathcal{N}_{x}(\mu_{1}, \Sigma _{1})$ and $\mathcal{N}_{x}(\mu_{2}, \Sigma _{2})$ can be formulated as:
\begin{equation}
  \alpha \mathcal{N}_{x}(\mu,\Sigma ) = \mathcal{N}_{x}(\mu_{1},\Sigma_{1})\mathcal{N}_{x}(\mu_{2},\Sigma_{2}),
\end{equation}
\begin{equation}
  \alpha = \mathcal{N}_{\mu1}(\mu2, \Sigma_{1} + \Sigma_{2}),
\end{equation}
where $\mu  = \mu_{1} + K(\mu_{2} - \mu_{1})$, $\Sigma = \Sigma_{1} - K\Sigma_{1}$, and $K$ is the Kalman gain, $K = \Sigma_{1}(\Sigma_{1} + \Sigma_{2})^{-1}$. To minimize the deviation of two rotated bounding box, we use a center
point loss $\mathcal{L}_{c}$ to narrow the distance between the center of the two Gaussian distributions. In this
way, $\alpha$ can be approximated as a constant, and the introduction of the $\mathcal{L}_{c}$ also allows the entire loss to continue to optimize the detector in non-overlapping cases. The overlap area can be calculated as follows:
\begin{equation}
  KFIoU=\frac{\mathcal{V}_{B3}(\Sigma )}{\mathcal{V}_{B1}(\Sigma_{1})+\mathcal{V}_{B2}(\Sigma _{2})-\mathcal{V}_{B3}(\Sigma )}.
\end{equation}
\begin{figure*}
  \includegraphics[width=\linewidth]{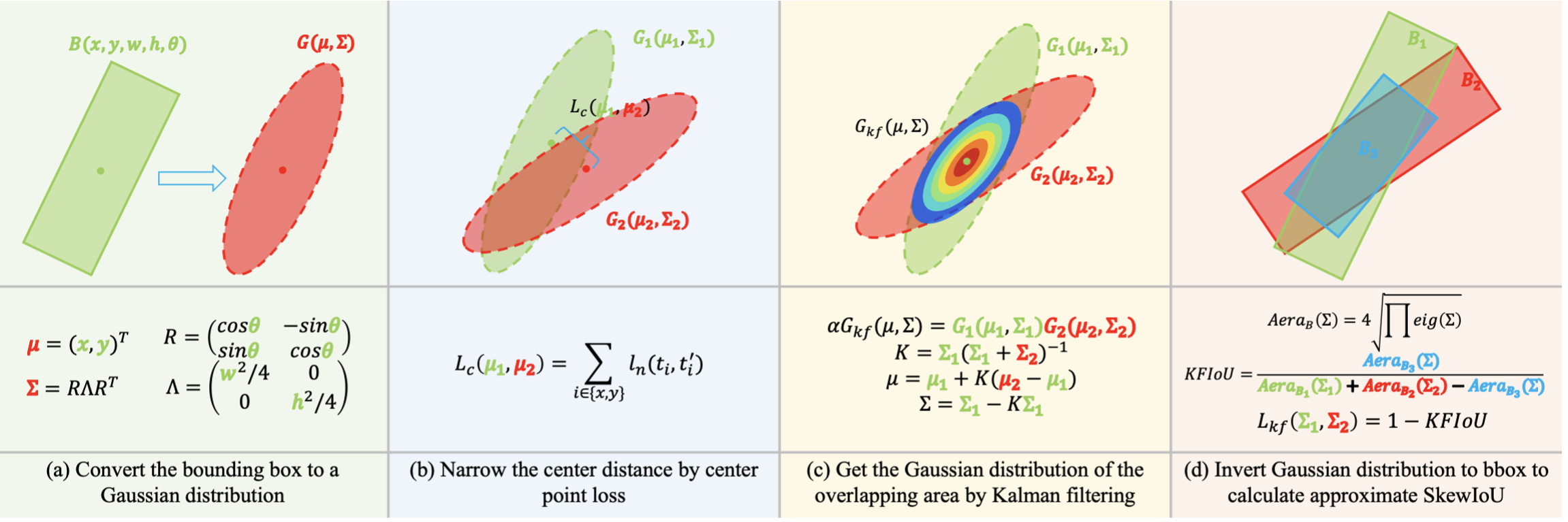}
  \caption{SkewIoU loss approximation process in two-dimensional space based on Gaussian product. Compared with GWD loss \cite{Yang2021RethinkingRO} and KLD loss \cite{KLD}, our approach follows the calculation process of SkewIoU without introducing additional hyperparameters. We believe such a design is more mathematically rigorous and more in line with SkewIoU loss}
  \label{}
\end{figure*}
where $B1$, $B2$ and $B3$ refer to the three different bounding boxes shown in the right part of Fig. 2, respectively. As we can see, the calculation process of $KFIoU$ is extremely complex.
After analyzing the metrics for bounding box regression of rotated object detection mentioned above, we found that geometric properties of rotated bounding box regression are actually not fully exploited in existing loss functions. Therefore, we propose $FPDIoU$ loss by minimizing the top-left, top-right, bottom-right and bottom-left points distance between predicted bounding box and the groundtruth bounding box for better training deep models of rotated object detection.
\section{Proposed Method}
It can be proved that four points can define a unique rotated rectangle in 2-D dimension. Inspired by the geometric properties of rotated rectangle, we designed a novel $IoU$-based metric named $FPDIoU$ to minimize the top-left, top-right, bottom-right and bottom-left points distance between the predicted bounding box and the groundtruth bounding box directly. In this way, we can improve the accuracy and efficiency of bounding box regression for rotated object detection.
\subsection{FPDIoU based on four points distance}

After analyzing the advantages and disadvantages of the $IoU$-based loss functions mentioned above, we start to think how to improve the accuracy and efficiency of bounding box regression. Generally speaking, we use the coordinates of top-left, top-right, bottom-right and bottom-left points to define a unique rotated rectangle. Inspired by the geometric properties of bounding boxes, we designed a novel $IoU$-based metric named $FPDIoU$ to minimize the top-left, top-right, bottom-right and bottom-left points distance between the predicted bounding box and the groundtruth bounding box directly. The calculation of $FPDIoU$ is summarized in Alg. \ref{alg1}.
\begin{algorithm}[h] %算法开始 
  \caption{Intersection over Union with Four Points Distance} %算法的题目 
  \label{alg1} %算法的标签 
  \begin{algorithmic}[1] %此处的[1]控制一下算法中的每句前面都有标号 
  \REQUIRE Two arbitrary convex shapes: $ A,B\subseteq \mathbb{S} \in \mathbb{R} ^{n}$, width and height of input image:$w,h$%输入条件(此处的REQUIRE默认关键字为Require，在上面已自定义为Input) 
  \ENSURE $FPDIoU(A,B)$ %输出结果(此处的ENSURE默认关键字为Ensure在上面已自定义为Output) 
% if-then-else 
\STATE For $A$ and $B$, $(x_{1}^{A},y_{1}^{A}),(x_{2}^{A},y_{2}^{A}),(x_{3}^{A},y_{3}^{A}),(x_{4}^{A},y_{4}^{A})$ denote the top-left, top-right, bottom-right and bottom-left point coordinates of $A$, respectively. $(x_{1}^{B},y_{1}^{B}),(x_{2}^{B},y_{2}^{B}),(x_{3}^{B},y_{3}^{B}),(x_{4}^{B},y_{4}^{B})$ denote the top-left, top-right, bottom-right and bottom-left point coordinates of $B$, respectively.
% \STATE For input image, we calculate the diagonal distance $d=\sqrt{(w^2+h^2)}$.
\STATE $A.sort(), B.sort()$

\STATE $d_{1}^{2}=(x_{1}^{B}-x_{1}^{A})^{2}+(y_{1}^{B}-y_{1}^{A})^{2}$ 
\STATE $d_{2}^{2}=(x_{2}^{B}-x_{2}^{A})^{2}+(y_{2}^{B}-y_{2}^{A})^{2}$
\STATE $d_{3}^{2}=(x_{3}^{B}-x_{3}^{A})^{2}+(y_{3}^{B}-y_{3}^{A})^{2}$ 
\STATE $d_{4}^{2}=(x_{4}^{B}-x_{4}^{A})^{2}+(y_{4}^{B}-y_{4}^{A})^{2}$
\STATE $FPDIoU(A,B)=\frac{A\bigcap B}{A\bigcup B}-\frac{d_{1}^{2}+d_{2}^{2}+d_{3}^{2}+d_{4}^{2}}{4*(w^2+h^2)}$
\end{algorithmic} 
\end{algorithm}

\begin{figure}
  \centering
  \includegraphics[width=\linewidth]{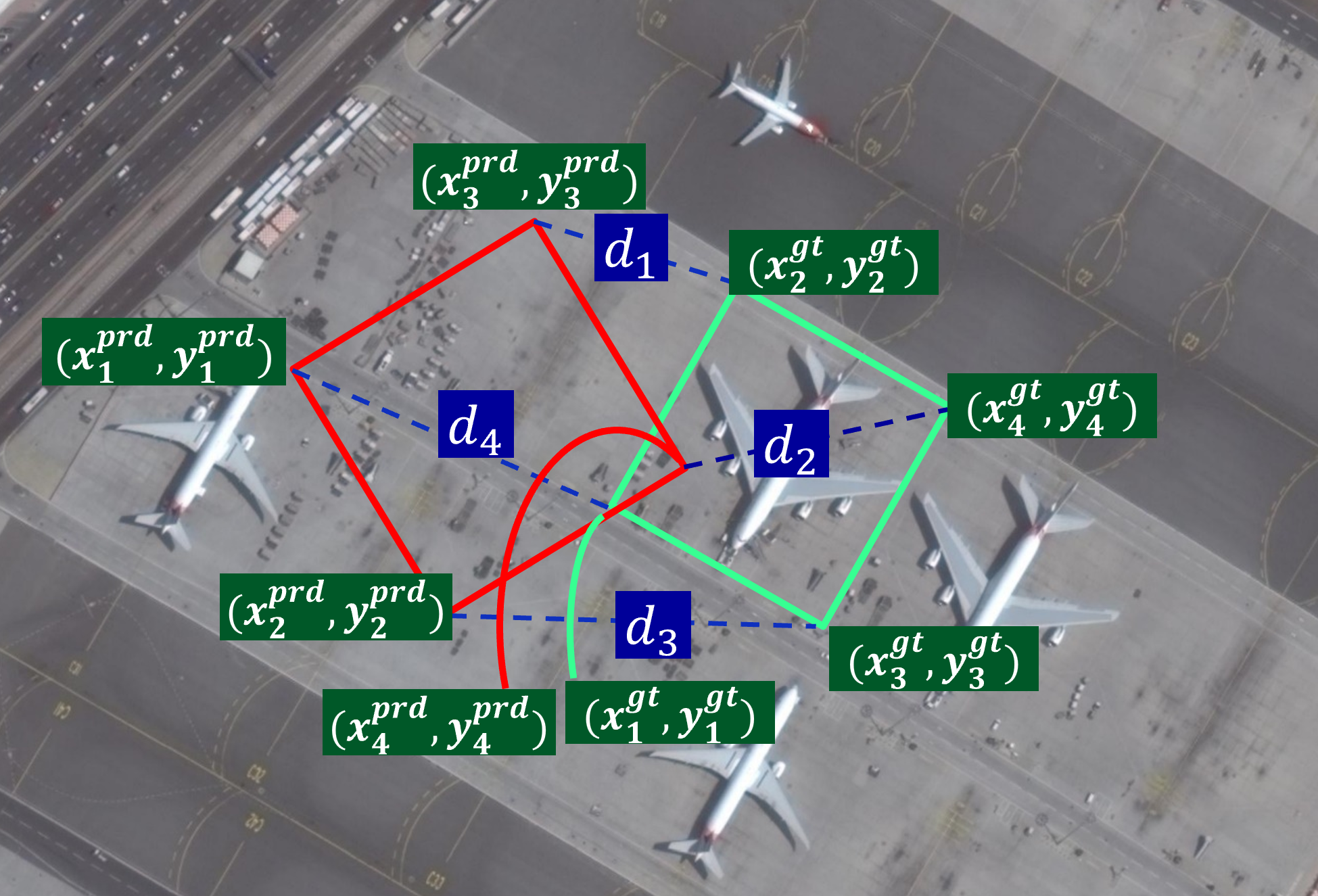}
  \caption{The regression process of our proposed $\mathcal{L}_{FPDIoU}$. The \textcolor{green}{green} box denotes the groundtruth bounding box and the \textcolor{red}{red} box denotes the predicted bounding box.}
  \label{FPDIoU}
\end{figure}

In order to ensure the one-to-one correspondence between the coordinates of the predicted bounding box and the groundtruth box no matter how the angle of rotation changes, we sort the coordinates of each bounding box in ascending order of $x$ value. Based on this operation, we can ensure $x_{1}^{A}<x_{2}^{A}\leq x_{3}^{A}<x_{4}^{A}$ and $x_{1}^{B}<x_{2}^{B}\leq x_{3}^{B}<x_{4}^{B}$. The calculation factors of $FPDIoU$ are shwon in Figure \ref{FPDIoU}.

In summary, our proposed $FPDIoU$ simplifies the similarity comparison between two bounding boxes, which can adapt to overlapping or nonoverlapping bounding box regression. Therefore, $FPDIoU$ can be a proper substitute for $IoU$ in all performance measures used in 2D computer vision tasks. In this paper, we only focus on 2D rotated object detection where we can easily derive an analytical solution for $FPDIoU$ to apply it as both metric and loss. The extension to non-axis aligned 3D cases is left as future work.
\subsection{The proposed FPDIoU loss}

In the training phase, each bounding box \\$\mathcal{B}_{prd} = [x^{prd},y^{prd},w^{prd},h^{prd},\theta^{prd}]^T$ predicted by the model is forced to approach its groundtruth box $\mathcal{B}_{gt} = [x^{gt},y^{gt},w^{gt},h^{gt},\theta^{gt}]^T$ by minimizing loss function below:

\begin{equation}
  \mathcal{L}=\underset{\Theta }{\min } \underset{\mathcal{B} _{gt}\in\mathbb{B}_{gt} }{\sum}\mathcal{L}(\mathcal{B}_{gt},\mathcal{B}_{prd}|\Theta),
\end{equation}
where $\mathbb{B}_{gt}$ is the set of ground-truth boxes, and $\Theta$ is the parameter of deep model for regression. A typical form of $\mathcal{L}$ is $\ell_{n}-norm$, for example, mean-square error (MSE) loss and Smooth-$\ell_{1}$ loss \cite{1992Robust}, which have been widely adopted in object detection \cite{bae2019object}; pedestrian detection \cite{brazil2017illuminating,zhou2019ssa}; scene text spotting \cite{huang2022swintextspotter,lyu2018mask}; 3D object detection \cite{zhou2018voxelnet,shi2019pointrcnn}; pose estimation \cite{sun2019deep,iskakov2019learnable}; and instance segmentation \cite{chen2019tensormask,Bolya_2019_ICCV}. However, recent researches suggest that $\ell_{n}$-norm-based loss functions are not consistent with the evaluation metric, that is, interaction over union (IoU), and instead propose $IoU$-based loss functions \cite{yu2016unitbox,tychsen2018improving,2019Generalized}.

\begin{algorithm}[h]%算法开始 
 
  \begin{algorithmic}[1] %此处的[1]控制一下算法中的每句前面都有标号 
  \REQUIRE The four points' coordinates of the predicted bounding box $\mathcal{B}_{prd}$ and groundtruth bounding box $\mathcal{B}_{gt}$: $\mathcal{B}_{prd}=(x_{1}^{prd},y_{1}^{prd},x_{2}^{prd},y_{2}^{prd},x_{3}^{prd},y_{3}^{prd},x_{4}^{prd},y_{4}^{prd})$,$\mathcal{B}_{gt}=(x_{1}^{gt},y_{1}^{gt},x_{2}^{gt},y_{2}^{gt},x_{3}^{gt},y_{3}^{gt},x_{4}^{gt},y_{4}^{gt})$, width and height of input image:$w,h$.
  \ENSURE $\mathcal{L}_{RotatedIoU}, \mathcal{L}_{FPDIoU}$
  % \mathcal{L}_{DIoU}, \mathcal{L}_{CIoU}, \mathcal{L}_{EIoU},  %输出结果(此处的ENSURE默认关键字为Ensure在上面已自定义为Output) 
% if-then-else 

\STATE $d_{1}^{2}=(x_{1}^{prd}-x_{1}^{gt})^{2}+(y_{1}^{prd}-y_{1}^{gt})^{2}$ 
\STATE $d_{2}^{2}=(x_{2}^{prd}-x_{2}^{gt})^{2}+(y_{2}^{prd}-y_{2}^{gt})^{2}$
\STATE $d_{3}^{2}=(x_{3}^{prd}-x_{3}^{gt})^{2}+(y_{3}^{prd}-y_{3}^{gt})^{2}$ 
\STATE $d_{4}^{2}=(x_{4}^{prd}-x_{4}^{gt})^{2}+(y_{4}^{prd}-y_{4}^{gt})^{2}$
\STATE Calculating rotated IoU of $\mathcal{B}_{gt}$ and $\mathcal{B}_{prd}$ as $RotatedIoU(\mathcal{B}_{gt}, \mathcal{B}_{prd})$.

\STATE $FPDIoU=RotatedIoU(\mathcal{B}_{gt}, \mathcal{B}_{prd})-\frac{d_{1}^{2}+d_{2}^{2}+d_{3}^{2}+d_{4}^{2}}{4*(h^2+w^2)}$
\STATE $\mathcal{L}_{RotatedIoU}=1-RotatedIoU$, $\mathcal{L}_{FPDIoU}=1-FPDIoU$.
% $\mathcal{L}_{GIoU}=1-GIoU$, $\mathcal{L}_{DIoU}=1-DIoU$, $\mathcal{L}_{CIoU}=1-CIoU$, $\mathcal{L}_{EIoU}=1-EIoU$, 
\end{algorithmic} 
\caption{IoU and FPDIoU as bounding box losses} %算法的题目 
\label{alg2} %算法的标签 
\end{algorithm}

Since back-propagating min, max and piece-wise linear functions, e.g. Relu, are feasible, it can be shown that every component in Alg. \ref{alg2} has a well-behaved derivative. Therefore, $RotatedIoU$, $GIoU$, $DIoU$, $PIoU$, $KFIoU$ and $FPDIoU$ can be directly used as loss functions, i.e. $\mathcal{L}_{RotatedIoU}$,$\mathcal{L}_{GIoU}$, $\mathcal{L}_{DIoU}$, $\mathcal{L}_{PIoU}$, $\mathcal{L}_{KFIoU}$ and $\mathcal{L}_{FPDIoU}$, for optimizing object detection or instance segmentation models. In this case, we are directly optimizing a metric as loss, which is an optimal choice for the metric. However, in all non-overlapping cases, $RotatedIoU$ has zero gradient, which affects both training accuracy and convergence speed. $FPDIoU$, in contrast, has a gradient in all possible cases, including non-overlapping situations. 

Considering the groundtruth bounding box, $\mathcal{B}_{gt}$ is a rotated rectangle with area bigger than zero, i.e. $A^{gt} > 0$. As a result, union area $\mathcal{U}>0$ for any predicted bounding box \\$\mathcal{B}_{prd}=(x_{1}^{prd},y_{1}^{prd},x_{2}^{prd},y_{2}^{prd},x_{3}^{prd},y_{3}^{prd},x_{4}^{prd},y_{4}^{prd})\in \mathbb{R}^{8}$. This ensures that the denominator in $IoU$ cannot be zero for any predicted value of outputs. In addition, for any values of $\mathcal{B}_{prd}=(x_{1}^{prd},y_{1}^{prd},x_{2}^{prd},y_{2}^{prd},x_{3}^{prd},y_{3}^{prd},x_{4}^{prd},y_{4}^{prd})\in \mathbb{R}^{8}$, the union area is always bigger than the intersection area, i.e. $ \mathcal{U} \geq\mathcal{I}$. As a result, $\mathcal{L}_{FPDIoU}$ is always bounded, i.e. $0\leq\mathcal{L}_{FPDIoU}< 2, \forall\mathcal{B}_{prd} \in \mathbb{R}^{8}$.

 We use the one-stage detector RTMDET and weakly-supervised detector H2RBox as the baseline. The rotated rectangle can be represented by $(x, y, w, h, \theta)$ or\\ $((x_{1},y_{1}), (x_{2},y_{2}), (x_{3},y_{3}), (x_{4},y_{4}))$. The two expressions mentioned above can be converted into each other without affecting the regression results. The relationship between them can be formulated as follows:
 \begin{equation}\label{xy}
  x=\frac{x_{2}+x_{1}}{2}, y=\frac{y_{2}+y_{1}}{2}
 \end{equation}
 \begin{equation}\label{wh}
  w=\sqrt{(x_{2}-x_{1})^2+(y_{2}-y_{1})^2}, h=\sqrt{(x_{3}-x_{1})^2+(y_{3}-y_{1})^2}
 \end{equation}
 \begin{equation}\label{theta}
  \theta=\arctan \frac{y_{2}-y_{3}}{x_{2}-x_{3}}
 \end{equation}
where $x, y, w, h$ denote the rotated bounding box's center coordinates, width and height, respectively. $\theta$ denotes the rotation angle of rotated bounding box. $(x_{1},y_{1}), (x_{2},y_{2}),$\\$(x_{3},y_{3}), (x_{4},y_{4})$ represent the top-left, top-right, bottom-right and bottom-left points coordinates, respectively. As Eq.\ref{xy}-\ref{theta} shows, all of the key factors of rotated bounding box can be represented by four points coordinates, which means our proposed $FPDIoU$ is not only considerate, but also simplifies the calculation process of rotated bounding box regression.

\section{Experiments}
\subsection{Datasets}
We choose some of the popular rotated object detection datasets to evaluate the performance of our proposed loss function. The details are shown as follow.\\
\textbf{DOTA \cite{DOTA}} is one of the largest datasets for oriented object detection in aerial images with three released versions: DOTA-v1.0, DOTA-v1.5 and DOTA-v2.0. DOTA-v1.0 contains 15 common categories, 2,806 images and 188,282 instances. The images of DOTA-v1.5 are the same as DOTA-v1.0, but some of the small instances (less than 10 pixels) are also annotated. Moreover, a new category, containing 402,089 instances in total is added in this version. While DOTA-v2.0 contains 18 common categories, 11,268 images and 1,793,658 instances. We divide the images into 600 *600 subimages with an overlap of 150 pixels and scale it to 800*800.\\
\textbf{HRSC2016 \cite{hrsc}} is a publicly available remote sensing image ship dataset contains images from two scenarios with ships on sea and close inshore. The training, validation and test set include 436, 181 and 444 images, covering 28 object classes.\\
\textbf{DIOR \cite{DIOR}} is a large-scale benchmark dataset for rotated object detection of optical remote sensing images. The dataset contains 23,463 images and 192,472 instances, covering 20 object classes. The 20 object classes are airplanes, airports, baseball fields, basketball courts, bridges, chimneys, dams, highway service areas, highway toll booths, ports, golf courses, surface athletic fields, flyovers, ships, stadiums, storage tanks, tennis courts, train stations, vehicles and wind mills.\\
\textbf{ICDAR 2017 RRC-MLT \cite{nayef2017icdar2017}} is a multilingual and multi-directional scene text detection and recognition dataset, which includes 9 languages including Chinese, Japanese, Korean, English, French, Arabic, Italian, German, and India. The scale of training, validation, and testing sets are 7200, 1800 and 9000, respectively.\\
\textbf{ICDAR 2019 RRC-MLT \cite{nayef2019icdar2019}} is a multilingual and multi-directional scene text detection and recognition dataset, including 10 languages such as Arabic, English, Japanese, Korean, and Chinese. The scale of training and testing sets are all 10 000.

\subsection{Experimental Settings}
The experimental environment of rotated object detectors described in this paper is as follows: the memory is 32GB, the operating system is windows 11, the CPU is Intel i9-12900k, and the graphics card is NVIDIA Geforce RTX 3090 with 24GB memory. In order to conduct a fair comparison, all of the experiments are implemented with PyTorch \cite{2017Automatic}. 

\subsection{Evaluation Protocol}
In this paper, we used mean Average Precision (mAP) over different class labels for a specific value of $IoU$ threshold in order to determine true positives and false positives. The main performance measure of rotated object detection used in our experiments is shown by precision and mAP@0.5:0.95. We report the mAP value for $IoU$ thresholds equal to 0.5 and 0.75, shown as AP50 and AP75 in the tables.
As for arbitrary orientation scene text spotting, the main performance measurements used in our experiments are shown by Precision, Recall and Hmean.

\subsection{Experimental Results}
\subsubsection{Comparison with the existing losses}
\begin{figure}[h]
  \centering
  \subfloat[HRSC2016]{
    \includegraphics[width=0.49\linewidth]{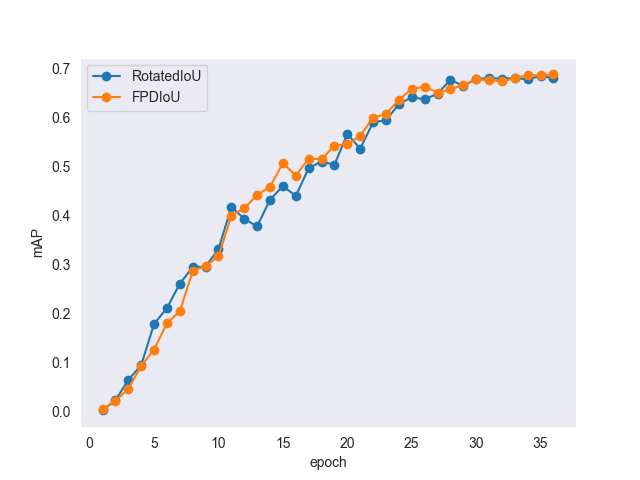}
  }
  \subfloat[DOTA]{
    \includegraphics[width=0.49\linewidth]{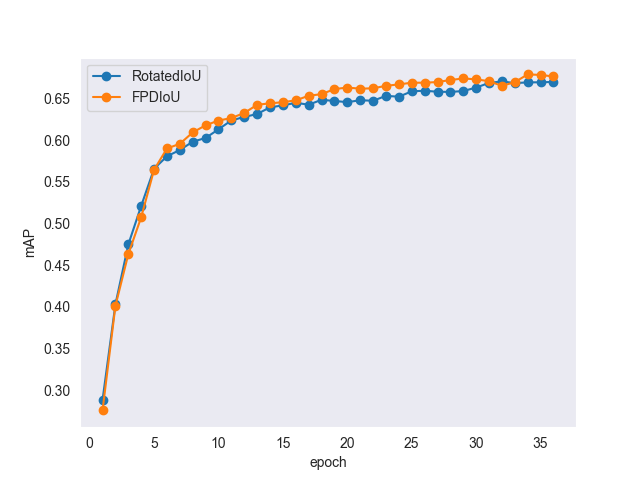}
  }
  \caption{Comparison of mAP value trends on HRSC2016 and DOTA test set training with RTMDET \cite{lyu2022rtmdet}.}
  \label{HRSC_train}
\end{figure}

\begin{figure}[h]
  \centering
  \subfloat[DIOR]{
    \includegraphics[width=0.49\linewidth]{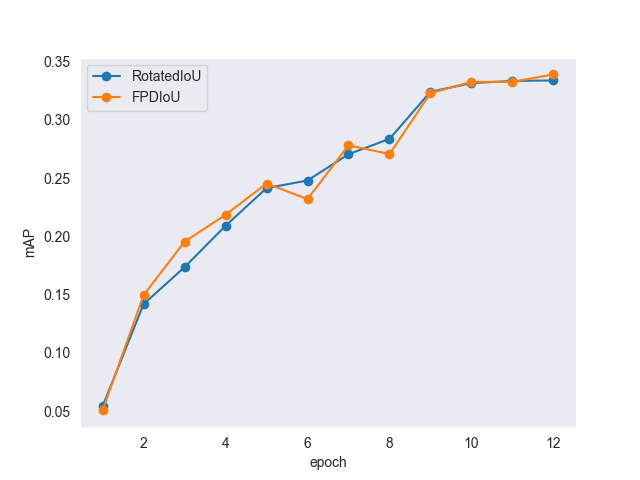}
  }
  \subfloat[DOTA]{
    \includegraphics[width=0.49\linewidth]{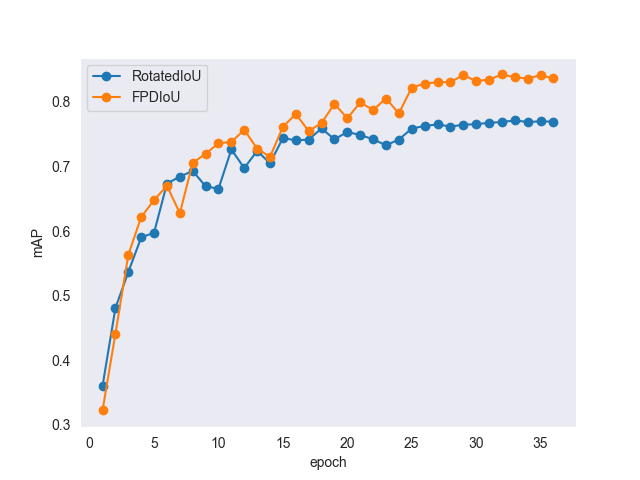}
  }
  \caption{Comparison of mAP value trends on DIOR and DOTA test set training with H2RBox \cite{yang2022h2rbox}.}
  \label{DIOR_train}
\end{figure}
We used the original CSPDarkNet implementation of RTMDET released by \cite{lyu2022rtmdet} and the official implementation of H2RBox to conduct our experiments. To train RTMDET and H2RBox using $RotatedIoU$, $GIoU$, $DIoU$,$PIoU$, $KFIoU$ and $FPDIoU$ losses, we simply replace the bounding box regression $RotatedIoU$ loss with $\mathcal{L}_{FPDIoU}$ loss explained in Alg. \ref{alg2}.\\
\textbf{Results of object detection for natural scenes}

In order to compare the regression process between $\mathcal{L}_{RotatedIoU}$ and $\mathcal{L}_{FPDIoU}$, we recorded the mAP values during the training process using $\mathcal{L}_{RotatedIoU}$ and $\mathcal{L}_{FPDIoU}$. As Figure \ref{HRSC_train} and Figure \ref{DIOR_train} shows, the model training with our proposed $\mathcal{L}_{FPDIoU}$ reached the best performance earlier than training with $\mathcal{L}_{RotatedIoU}$, which means our proposed $\mathcal{L}_{FPDIoU}$ has faster convergence speed and higher accuracy on rotated object detection tasks.
\begin{table}[h]
  \caption{Comparison between the performance of H2RBox \cite{yang2022h2rbox} trained using its own loss ($\mathcal{L}_{RotatedIoU}$) as well as $\mathcal{L}_{GIoU}$, $\mathcal{L}_{DIoU}$, $\mathcal{L}_{PIoU}$, $\mathcal{L}_{KFIoU}$ and $\mathcal{L}_{FPDIoU}$. The results are reported on the \textbf{test set of DOTA-v1.0}.}
  \label{h2rboxDOTA}
  \centering
  \small
  \setlength{\tabcolsep}{3pt}
  \begin{tabular}{c|cccc}
  \toprule
  \diagbox{Loss}{Evaluation}&
  mAP&AP50&AP75&speed(iter/s)\\
  \midrule
    $\mathcal{L}_{RotatedIoU}$& 
    30.84&63&25.83&\textbf{1.11}
    \\
    \hline
    $\mathcal{L}_{GIoU}$& 
    31.94&64&26.93&1.05\\
    Relative improv(\%)&3.56&1.58&4.25&-5.4\\
    \hline
    $\mathcal{L}_{DIoU}$& 
    32.08&65.8&27.63&0.99
    \\
    Relative improv(\%)&4.02&4.4&6.96&-10.8\\
    \hline
    $\mathcal{L}_{PIoU}$& 
    33.28&66.6&28.23&0.91
    \\
    Relative improv(\%)&7.91&5.71&9.29&-18.01\\
    \hline
    $\mathcal{L}_{KFIoU}$& 
    35.18&67.2&27.63&0.89
    \\
    Relative improv(\%)&14.07&4.4&6.96&-10.8\\
    \hline
    $\mathcal{L}_{FPDIoU}$&\textbf{40.13}
    &\textbf{71.74}&\textbf{38.41}&0.73\\
    Relative improv(\%)&30.12&13.87&48.7&-34.2\\
    \bottomrule
  \end{tabular}
  
\end{table}

\begin{table}[h]
  \caption{Comparison between the performance of H2RBox \cite{yang2022h2rbox} trained using its own loss ($\mathcal{L}_{RotatedIoU}$) as well as $\mathcal{L}_{GIoU}$, $\mathcal{L}_{DIoU}$, $\mathcal{L}_{PIoU}$, $\mathcal{L}_{KFIoU}$ and $\mathcal{L}_{FPDIoU}$. The results are reported on the \textbf{test set of DIOR}.}
  \label{h2rboxDIOR}
  \centering
  \small
  \setlength{\tabcolsep}{3pt}
  \begin{tabular}{c|cccc}
  \toprule
  \diagbox{Loss}{Evaluation}&
  mAP&AP50&AP75&speed(iter/s)\\
  \midrule
    $\mathcal{L}_{RotatedIoU}$&33.39
    &56.4&33.4&3.9
    \\
    \hline
    $\mathcal{L}_{GIoU}$&33.56
    &56.7&33.45&3.75
    \\
    Relative improv(\%)&0.51&0.53&0.14&-5.4\\
    
    \hline
    $\mathcal{L}_{DIoU}$&33.69&57.1&33.49&3.85
    \\
    Relative improv(\%)&0.89&1.24&0.26&-1.28\\
    \hline
    $\mathcal{L}_{PIoU}$& 
    33.78&57.3&33.49&3.88
    \\
    Relative improv(\%)&1.16&1.59&0.26&-0.51\\
    \hline
    $\mathcal{L}_{KFIoU}$& 
    33.8&57.6&33.5&4.01
    \\
    Relative improv(\%)&1.22&2.12&0.29&2.82\\
    \hline
    $\mathcal{L}_{FPDIoU}$&\textbf{33.9}
    &\textbf{57.7}&\textbf{33.5}&\textbf{4.17}\\
    Relative improv(\%)&1.52&2.3&0.29&6.9\\
    \bottomrule
  \end{tabular}
\end{table}

To make more intuitive comparison of the performance between the existing $IOU$-based loss functions and $\mathcal{L}_{FPDIoU}$ proposed by us, we conduct further experiments using $\mathcal{L}_{RotatedIoU}$, $\mathcal{L}_{GIoU}$, $\mathcal{L}_{DIoU}$, $\mathcal{L}_{PIoU}$, $\mathcal{L}_{KFIoU}$ and $\mathcal{L}_{FPDIoU}$ to train the SOTA rotated object detectors RTMDET\cite{lyu2022rtmdet} and H2RBox\cite{yang2022h2rbox}. We give the comparison results from two aspects including accuracy and efficiency. The experimental results are shown as Table \ref{h2rboxDOTA}-\ref{rtmdetHRSC}.

\begin{table}[h]
  \caption{Comparison between the performance of RTMDET \cite{lyu2022rtmdet} trained using its own loss ($\mathcal{L}_{RotatedIoU}$) as well as $\mathcal{L}_{GIoU}$, $\mathcal{L}_{DIoU}$, $\mathcal{L}_{PIoU}$, $\mathcal{L}_{KFIoU}$ and $\mathcal{L}_{FPDIoU}$. The results are reported on the \textbf{test set of DOTA-v1.0}.}
  \label{rtmdetDOTA}
  \centering
  \small
  \setlength{\tabcolsep}{3pt}
  \begin{tabular}{c|cccc}
  \toprule
  \diagbox{Loss}{Evaluation}&
  mAP&AP50&AP75&speed(iter/s)\\
  \midrule
    $\mathcal{L}_{RotatedIoU}$&46.16
    &74.94&49.06&2.73
    \\
    \hline
    $\mathcal{L}_{GIoU}$& 
    46.33&75.03&49.2&2.33\\
    Relative improv(\%)&0.36&0.12&0.28&-14.65\\
    \hline
    $\mathcal{L}_{DIoU}$& 
    46.73&75.2&49.55&2.05
    \\
    Relative improv(\%)&1.23&0.34&0.99&-24.9\\
    \hline
    $\mathcal{L}_{PIoU}$& 
    47.1&75.28&48.23&0.91
    \\
    Relative improv(\%)&7.91&5.71&9.29&-18.01\\
    \hline
    $\mathcal{L}_{KFIoU}$& 
    47.28&75.38&49.63&0.89
    \\
    Relative improv(\%)&2.42&0.58&1.16&-10.8\\
    \hline
    $\mathcal{L}_{FPDIoU}$&\textbf{47.37}
    &\textbf{75.38}&\textbf{50.71}&\textbf{3.75}\\
    Relative improv(\%)&2.62&0.58&3.36&37.36\\
    \bottomrule
  \end{tabular}
  
\end{table}

\begin{table}[h]
  \caption{Comparison between the performance of RTMDET \cite{lyu2022rtmdet} trained using its own loss ($\mathcal{L}_{RotatedIoU}$) as well as $\mathcal{L}_{GIoU}$, $\mathcal{L}_{DIoU}$, $\mathcal{L}_{PIoU}$, $\mathcal{L}_{KFIoU}$ and $\mathcal{L}_{FPDIoU}$. The results are reported on the \textbf{test set of HRSC2016}.}
  \label{rtmdetHRSC}
  \centering
  \small
  \setlength{\tabcolsep}{3pt}
  \begin{tabular}{c|cccc}
  \toprule
  \diagbox{Loss}{Evaluation}&
  mAP&AP50&AP75&speed(iter/s)\\
  \midrule
    $\mathcal{L}_{RotatedIoU}$&65.69
    &70.3&61.1&54.94
    \\
    \hline
    $\mathcal{L}_{GIoU}$& 
    65.77&70.33&61.2&62.3\\
    Relative improv(\%)&0.12&0.04&0.16&13.39\\
    \hline
    $\mathcal{L}_{DIoU}$& 
    65.97&70.35&61.6&66.2
    \\
    Relative improv(\%)&0.42&0.07&0.81&20.49\\
    \hline
    $\mathcal{L}_{PIoU}$& 
    66.1&70.35&61.85&69.5
    \\
    Relative improv(\%)&0.62&0.07&1.22&26.5\\
    \hline
    $\mathcal{L}_{KFIoU}$& 
    66.25&70.38&62.1&73.3
    \\
    Relative improv(\%)&0.85&0.11&1.63&33.41\\
    \hline
    $\mathcal{L}_{FPDIoU}$&\textbf{66.37}
    &\textbf{70.4}&\textbf{62.3}&\textbf{81.96}\\
    Relative improv(\%)&1.03&0.14&1.96&49.18\\
    \bottomrule
  \end{tabular}
\end{table}

\begin{table}[h]
  \caption{Comparison between the performance of RTMDET \cite{lyu2022rtmdet} trained using its own loss ($\mathcal{L}_{RotatedIoU}$) as well as $\mathcal{L}_{GIoU}$, $\mathcal{L}_{DIoU}$, $\mathcal{L}_{PIoU}$, $\mathcal{L}_{KFIoU}$ and $\mathcal{L}_{FPDIoU}$. The results are reported on the \textbf{test set of ICDAR 2017 RRC-MLT}.}
  \label{rtmdetIC17}
  \centering
  \small
  \setlength{\tabcolsep}{3pt}
  \begin{tabular}{c|cccc}
  \toprule
  \diagbox{Loss}{Evaluation}&
  P&R&H&speed(iter/s)\\
  \midrule
    $\mathcal{L}_{RotatedIoU}$& 
    82.55&54.77&65.85&2.29\\
    \hline
    $\mathcal{L}_{GIoU}$& 
    82.94&\textbf{54.87}&65.93&2.32\\
    Relative improv(\%)&0.47&0.18&0.12&1.31\\
    \hline
    $\mathcal{L}_{DIoU}$& 
    83.08&54.1&65.98&2.39
    \\
    Relative improv(\%)&0.64&-1.2&0.19&4.36\\
    \hline
    $\mathcal{L}_{PIoU}$& 
    83.28&54.26&66.03&2.43
    \\
    Relative improv(\%)&0.88&-0.93&0.27&6.11\\
    \hline
    $\mathcal{L}_{KFIoU}$&
    83.98&54.29&65.96&2.89\\
    Relative improv(\%)&1.73&-0.87&0.16&26.2\\
    \hline
    $\mathcal{L}_{FPDIoU}$&\textbf{84.19}&54.33&\textbf{66.04}&\textbf{2.93}\\
    Relative improv(\%)&1.98&-0.8&0.28&27.94\\
    \bottomrule
  \end{tabular}
\end{table}
\begin{table}[h]
  \caption{Comparison between the performance of RTMDET \cite{lyu2022rtmdet} trained using its own loss ($\mathcal{L}_{RotatedIoU}$) as well as $\mathcal{L}_{GIoU}$, $\mathcal{L}_{DIoU}$, $\mathcal{L}_{PIoU}$, $\mathcal{L}_{KFIoU}$ and $\mathcal{L}_{FPDIoU}$. The results are reported on the \textbf{test set of ICDAR 2019 RRC-MLT}.}
  \label{rtmdetIC19}
  \centering
  \small
  \setlength{\tabcolsep}{3pt}
  \begin{tabular}{c|cccc}
  \toprule
  \diagbox{Loss}{Evaluation}&
  P&R&H&speed(iter/s)\\
  \midrule
    $\mathcal{L}_{RotatedIoU}$& 
    83.20&55.63&66.68&2.29\\
    \hline
    $\mathcal{L}_{GIoU}$& 
    83.39&55.2&66.72&2.22\\
    Relative improv(\%)&3.56&1.58&4.25&-5.4\\
    \hline
    $\mathcal{L}_{DIoU}$& 
    83.60&55.33&66.75&2.39
    \\
    Relative improv(\%)&0.48&0.53&0.10&4.36\\
    \hline
    $\mathcal{L}_{PIoU}$& 
    83.68&55.6&66.83&2.41
    \\
    Relative improv(\%)&0.57&0.05&0.22&5.24\\
    \hline
    $\mathcal{L}_{KFIoU}$&
    84.18&\textbf{55.72}&66.86&2.41\\
    Relative improv(\%)&1.17&0.16&0.26&5.24\\
    \hline
    $\mathcal{L}_{FPDIoU}$&\textbf{84.84}&55.21&\textbf{66.89}&\textbf{2.43}\\
    Relative improv(\%)&1.97&0.75&0.31&6.11\\
    \bottomrule
  \end{tabular}
\end{table}
As we can see, no matter which rotated object detectors we use, the models training with our proposed $\mathcal{L}_{FPDIoU}$ achieve better performance than using existing losses for rotated object detection. In most of the cases, the $mAP$ values and inference speed increased obviously using $\mathcal{L}_{FPDIoU}$.

\begin{table*}[b]
  \caption{Rotated object detection results test on DOTA-v1.0 test dataset. \textcolor{red}{Red} and \textcolor{blue}{blue}: top two performances }
  \label{sota-compare}
  \centering
  \footnotesize
  \setlength{\tabcolsep}{3pt}
  \begin{tabular}{c|c|ccccccccccccccc|c}
  \toprule
  &Method&PL&BD&BR&GTF&SV&LV&SH&TC&BC&ST&SBF&RA&HA&SP&HC&$mAP_{50}$\\
  \midrule
  \multirow{10}*{\rotatebox{90}{Single-stage}}
  &PIoU\cite{PIoU}&80.90&69.70&24.10&60.20&38.30&64.40&64.80&90.90&77.20&70.40&46.50&37.10&57.10&61.90&64.00&60.50\\
  &$O^{2}$-DNet\cite{o2DNet}&89.31&82.14&47.33&61.21&71.32&74.03&78.62&90.76&82.23&81.36&60.93&60.17&58.21&66.98&61.03&71.04\\
  &DAL\cite{ming2021dynamic}&88.61&79.69&46.27&70.37&65.89&76.10&78.53&90.84&79.98&78.41&58.71&62.02&69.23&71.32&60.65&71.78\\
  &P-RSDet\cite{zhou2020objects}&88.58&77.83&50.44&69.29&71.10&75.79&78.66&90.88&80.10&81.71&57.92&63.03&66.30&69.77&63.13&72.30\\
  &BBAVectors\cite{yi2021oriented}&88.35&79.96&50.69&62.18&78.43&78.98&87.94&90.85&83.58&84.35&54.13&60.24&65.22&64.28&55.70&72.32\\
  &DRN\cite{DRN}&89.71&82.34&47.22&64.10&76.22&74.43&85.84&90.57&86.18&84.89&57.65&61.93&69.30&69.63&58.48&73.23\\
  &DCL\cite{DCL}&89.10&84.13&50.15&73.57&71.48&58.13&78.00&90.89&86.64&86.78&67.97&67.25&65.63&74.06&67.05&74.06\\
  &PolarDet\cite{PolarDet}&89.65&87.07&48.14&70.97&78.53&80.34&87.45&90.76&85.63&86.87&61.64&70.32&71.92&73.09&67.15&76.64\\
  &GWD\cite{Yang2021RethinkingRO}&86.96&83.88&54.36&77.53&74.41&68.48&80.34&86.62&83.41&85.55&73.47&67.77&72.57&75.76&73.40&76.30\\
  &KFIoU\cite{yang2022kfiou}&89.46&85.72&54.94&80.37&77.16&69.23&80.90&90.79&87.79&86.13&73.32&68.11&75.23&71.61&69.49&77.35\\
  \midrule
  \multirow{11}*{\rotatebox{90}{Refine-stage}}
  &CFC-Net\cite{CFCNet}&89.08&80.41&52.41&70.02&76.28&78.11&87.21&90.89&84.47&85.64&60.51&61.52&67.82&68.02&50.09&73.50\\
  &R3Det\cite{yang2021r3det}&89.80&83.77&48.11&66.77&78.76&83.27&87.84&90.82&85.38&85.51&65.67&62.68&67.53&78.56&72.62&76.47\\
  &CFA\cite{CFA}&89.08&83.20&54.37&66.87&81.23&80.96&87.17&90.21&84.32&86.09&52.34&69.94&75.52&80.76&67.96&76.67\\
  &DAL\cite{ming2021dynamic}&89.69&83.11&55.03&71.00&78.30&81.90&88.46&90.89&84.97&87.46&64.41&65.65&76.86&72.09&64.35&76.95\\
  &DCL\cite{DCL}&89.26&83.60&53.54&72.76&79.04&82.56&87.31&90.67&86.59&86.98&67.49&66.88&73.29&70.56&69.99&77.37\\
  &RIDet\cite{RIDet}&89.31&80.77&54.07&76.38&79.81&81.99&89.13&90.72&83.58&87.22&64.42&67.56&78.08&79.17&62.07&77.62\\
 &S2A-Net\cite{s2anet}&88.89&83.60&57.74&81.95&79.94&83.19&89.11&90.78&84.87&87.81&70.30&68.25&78.30&77.01&69.58&79.42\\
  &R3Det-GWD\cite{Yang2021RethinkingRO}&89.66&84.99&59.26&82.19&78.97&84.83&87.70&90.21&86.54&86.85&73.47&67.77&76.92&79.22&74.92&80.23\\
 &R3Det-KLD\cite{KLD}&89.92&85.13&59.19&81.33&78.82&84.38&87.50&89.80&87.33&87.00&72.57&71.35&77.12&79.34&78.68&80.63\\
  &R3Det-KFIoU\cite{yang2022kfiou}&88.89&85.14&60.05&81.13&81.78&85.71&88.27&90.87&87.12&87.91&69.77&73.70&79.25&81.31&74.56&\textcolor{blue}{81.03}\\

 \midrule
 \multirow{11}*{\rotatebox{90}{Two-stage}}
  &ICN\cite{ICN}&81.40&74.30&47.70&70.30&64.90&67.80&70.00&90.80&79.10&78.20&53.60&62.90&67.00&64.20&50.20&68.20\\
  &RoI-Trans\cite{RoI-trans}&88.64&78.52&43.44&75.92&68.81&73.68&83.59&90.74&77.27&81.46&58.39&53.54&62.83&58.93&47.67&69.56\\
  &SCRDet\cite{SCRDet}&89.98&80.65&52.09&68.36&68.36&60.32&72.41&90.85&87.94&86.86&65.02&66.68&66.25&68.24&65.21&72.61\\
  &Gliding Vertex\cite{GlidingVertex}&89.64&85.00&52.26&77.34&73.01&73.14&86.82&90.74&79.02&86.81&59.55&70.91&72.94&70.86&57.32&75.02\\
  &Mask OBB\cite{Wang2019MaskOA}&89.56&85.95&54.21&72.90&76.52&74.16&85.63&89.85&83.81&86.48&54.89&69.64&73.94&69.06&63.32&75.33\\
  &CenterMap\cite{CenterMap}&89.83&84.41&54.60&70.25&77.66&78.32&87.19&90.66&84.89&85.27&56.46&69.23&74.13&71.56&66.06&76.03\\
  &CSL\cite{CSL}&90.25&85.53&54.64&75.31&70.44&73.51&77.62&90.84&86.15&86.69&69.60&68.04&73.83&71.10&68.93&76.17\\
  &RSDet-II\cite{RSDet++}&89.93&84.45&53.77&74.35&71.52&78.31&78.12&91.14&87.35&86.93&65.64&65.17&75.35&79.74&63.31&76.34\\
  &SCRDet++\cite{SCRDet++}&90.05&84.39&55.44&73.99&77.54&71.11&86.05&90.67&87.32&87.08&69.62&68.90&73.74&71.29&65.08&76.81\\
  &ReDet\cite{ReDet}&88.81&82.48&60.83&80.82&78.34&86.06&88.31&90.87&88.77&87.03&68.65&66.90&79.26&79.71&74.67&80.10\\
  &Oriented R-CNN\cite{OrientedR-CNN}&89.84&85.43&61.09&79.82&79.71&85.35&88.82&90.88&86.68&87.73&72.21&70.80&82.42&78.18&74.11&80.87\\
  &RoI-Trans-KFIoU\cite{yang2022kfiou}&89.44&84.41&62.22&82.51&80.10&86.07&88.68&90.90&87.32&88.38&72.80&71.95&78.96&74.95&75.27&80.93\\
  &\textbf{RTMDET-FPDIoU(Ours)}&89.34&84.41&60.22&78.18&81.50&85.30&88.84&90.87&86.41&87.55&72.70&71.95&78.96&80.31&75.27&\textcolor{red}{81.33}\\
    \bottomrule
  \end{tabular}
\end{table*}

\begin{figure*}[ht]
  \subfloat[Rotated object detection results training with $\mathcal{L}_{RotatedIoU}$]{\includegraphics[width=0.25\linewidth,height=0.18\linewidth]{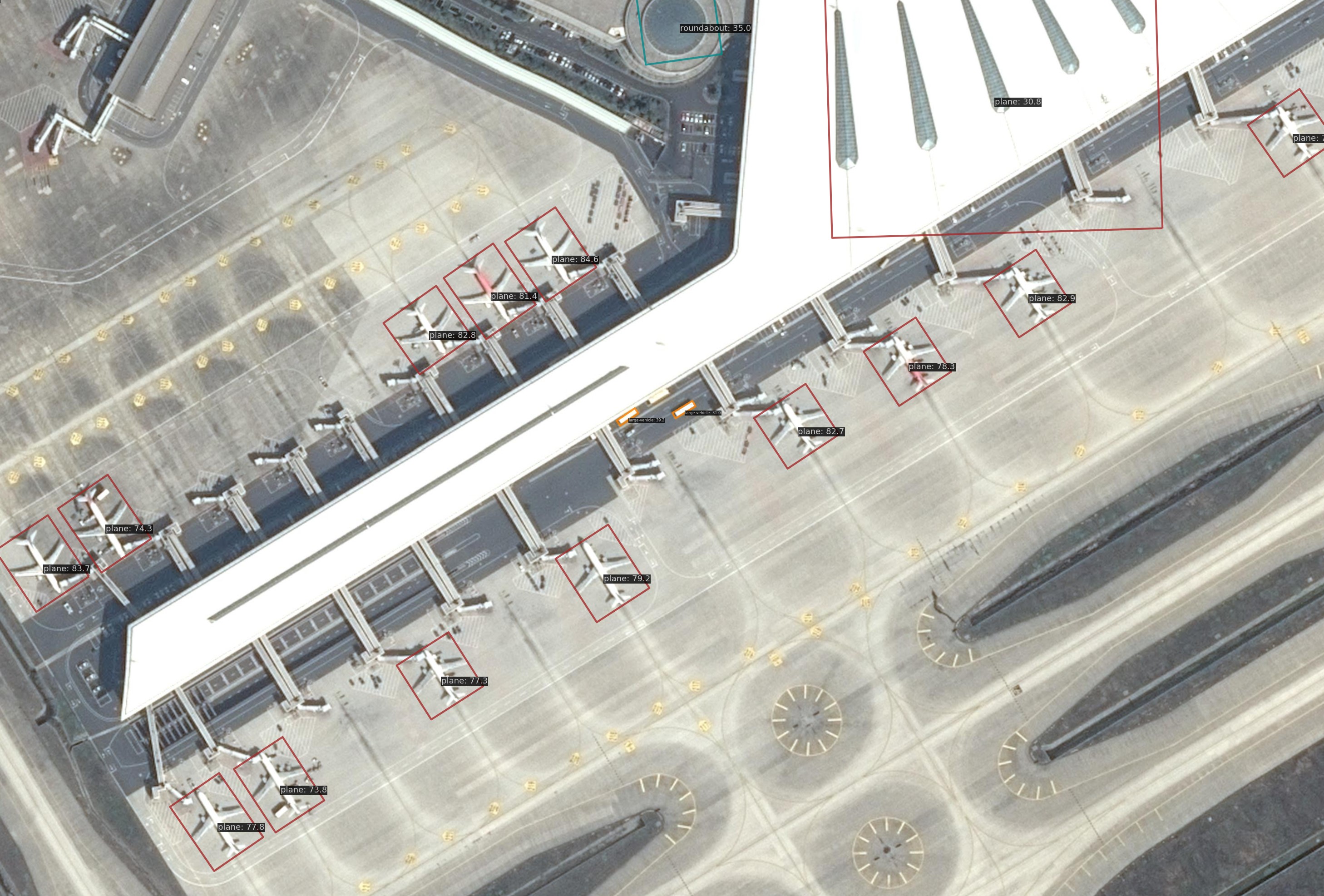}
  \includegraphics[width=0.25\linewidth,height=0.18\linewidth]{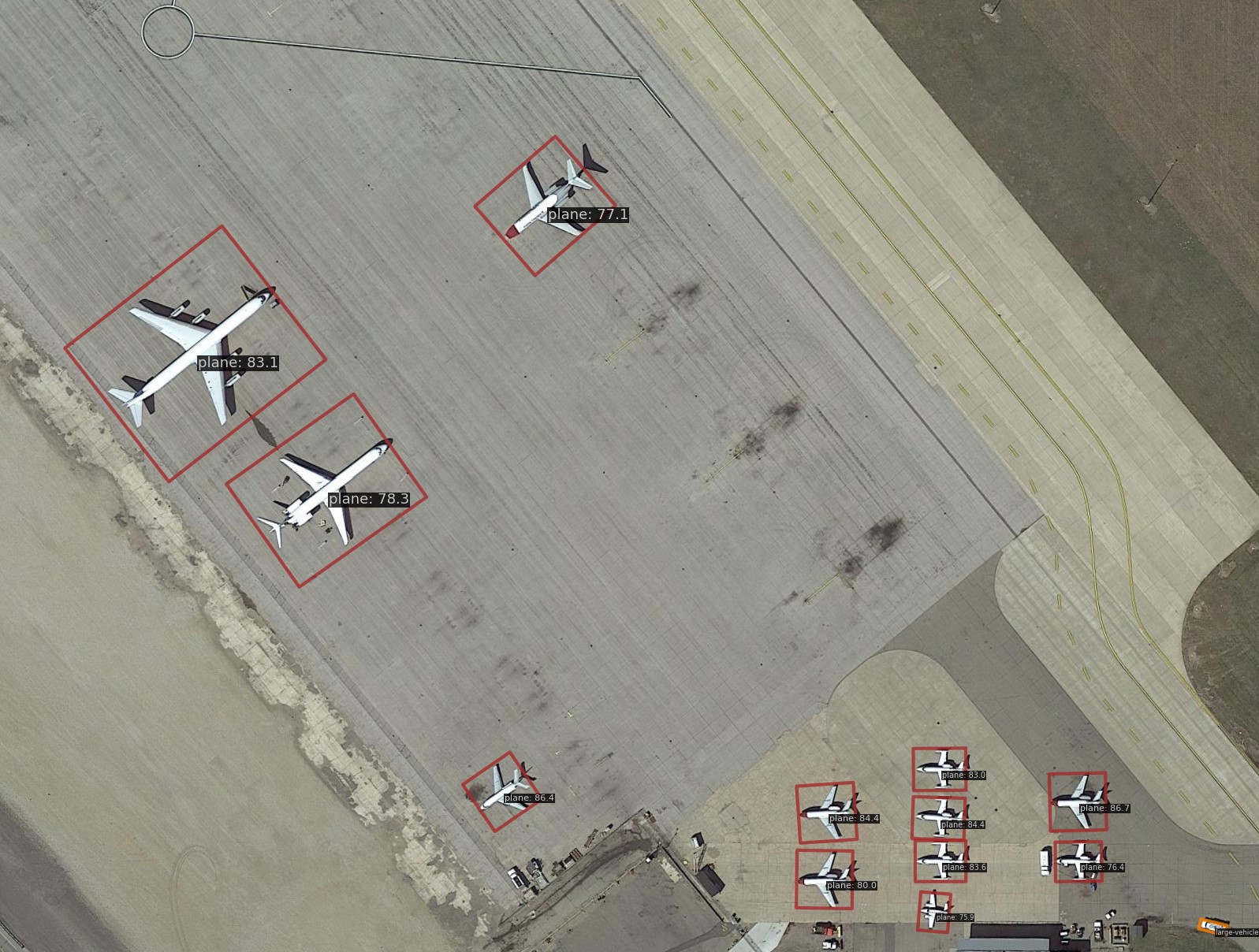}
  \includegraphics[width=0.25\linewidth,height=0.18\linewidth]{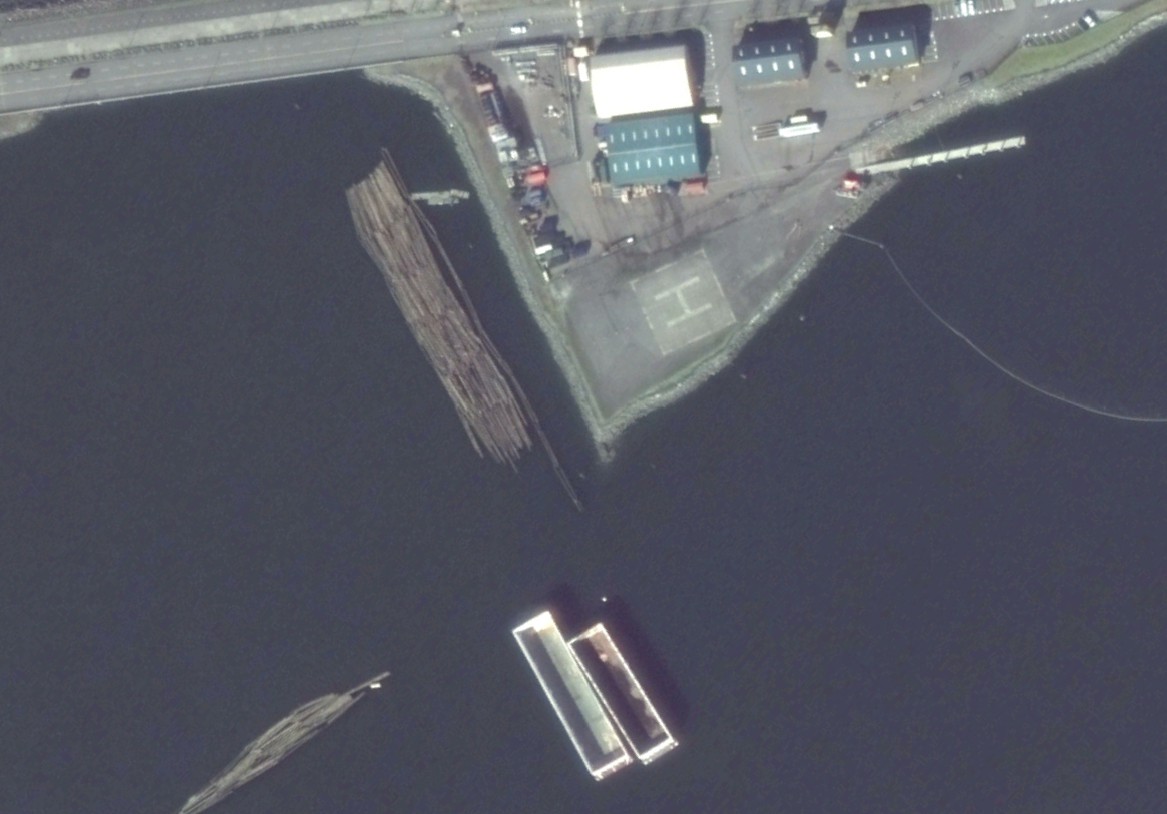}
  \includegraphics[width=0.25\linewidth,height=0.18\linewidth]{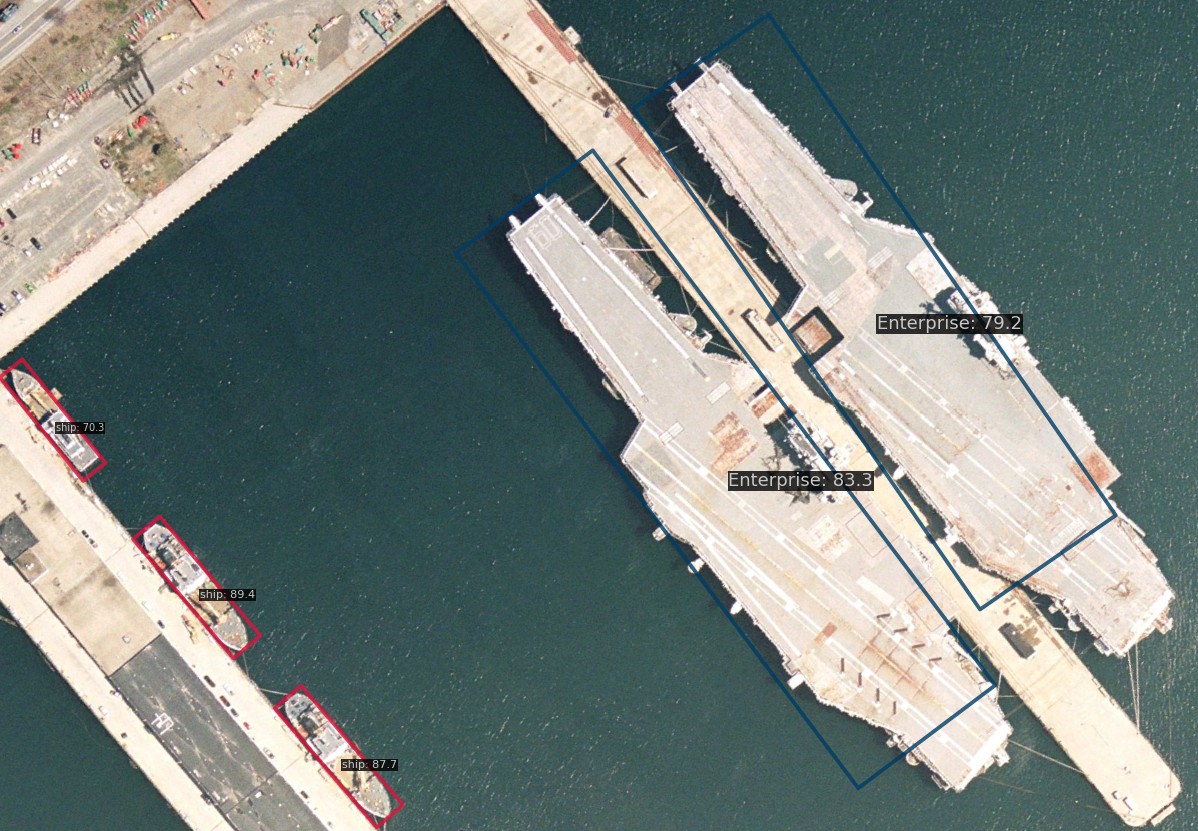}
  
  }\\
  \subfloat[Rotated object detection results training with $\mathcal{L}_{FPDIoU}$]{\includegraphics[width=0.25\linewidth,height=0.18\linewidth]{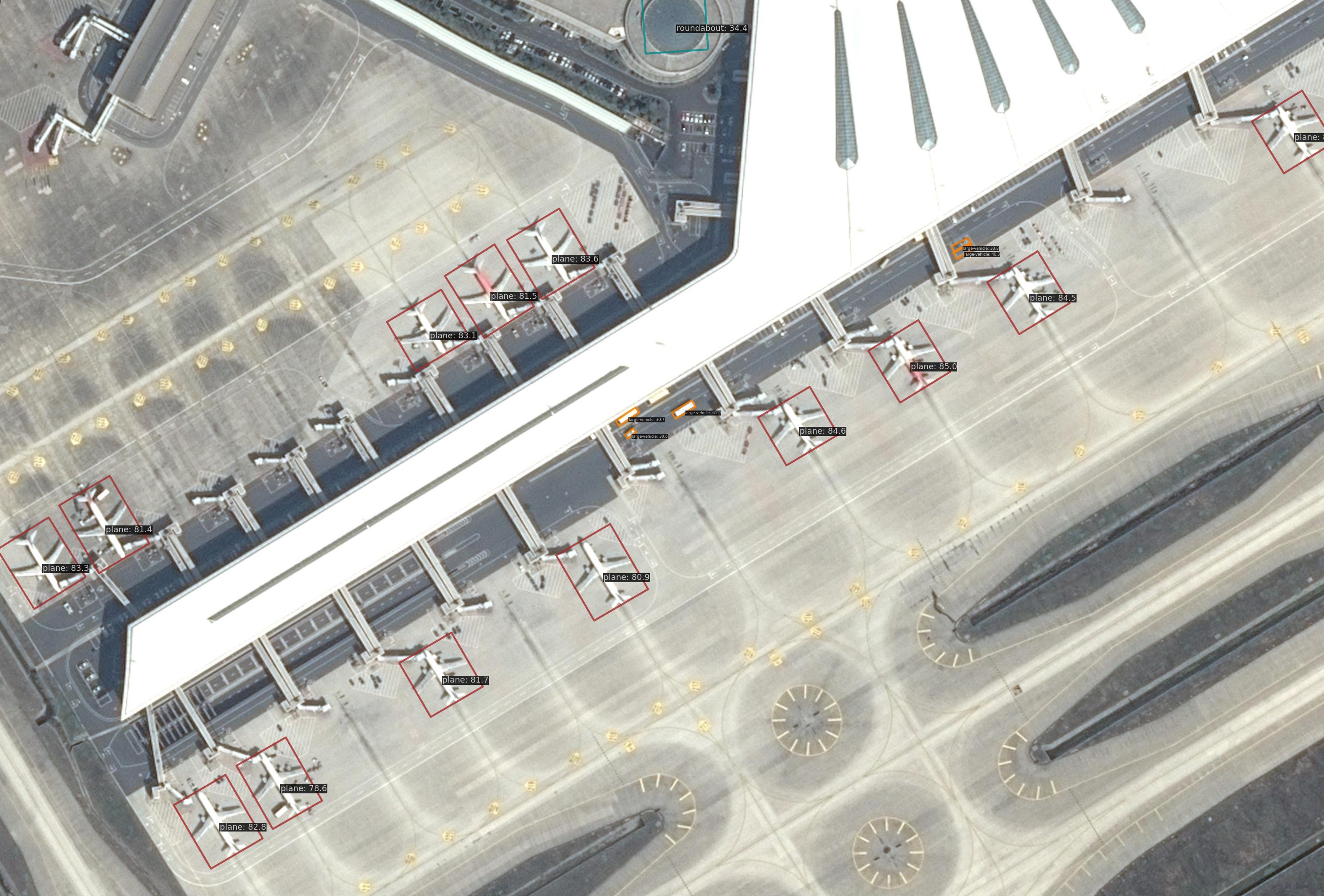}
  \includegraphics[width=0.25\linewidth,height=0.18\linewidth]{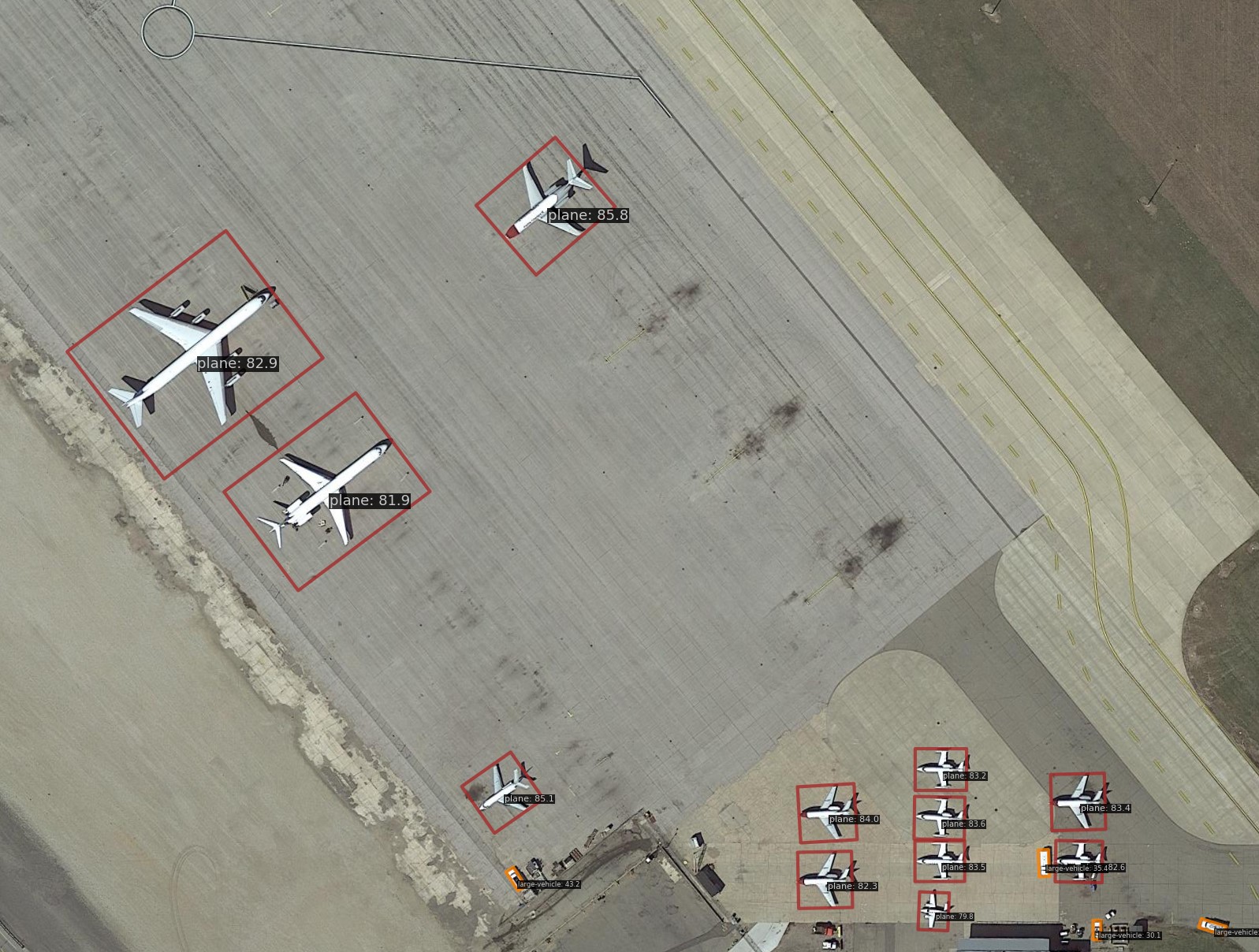}
  \includegraphics[width=0.25\linewidth,height=0.18\linewidth]{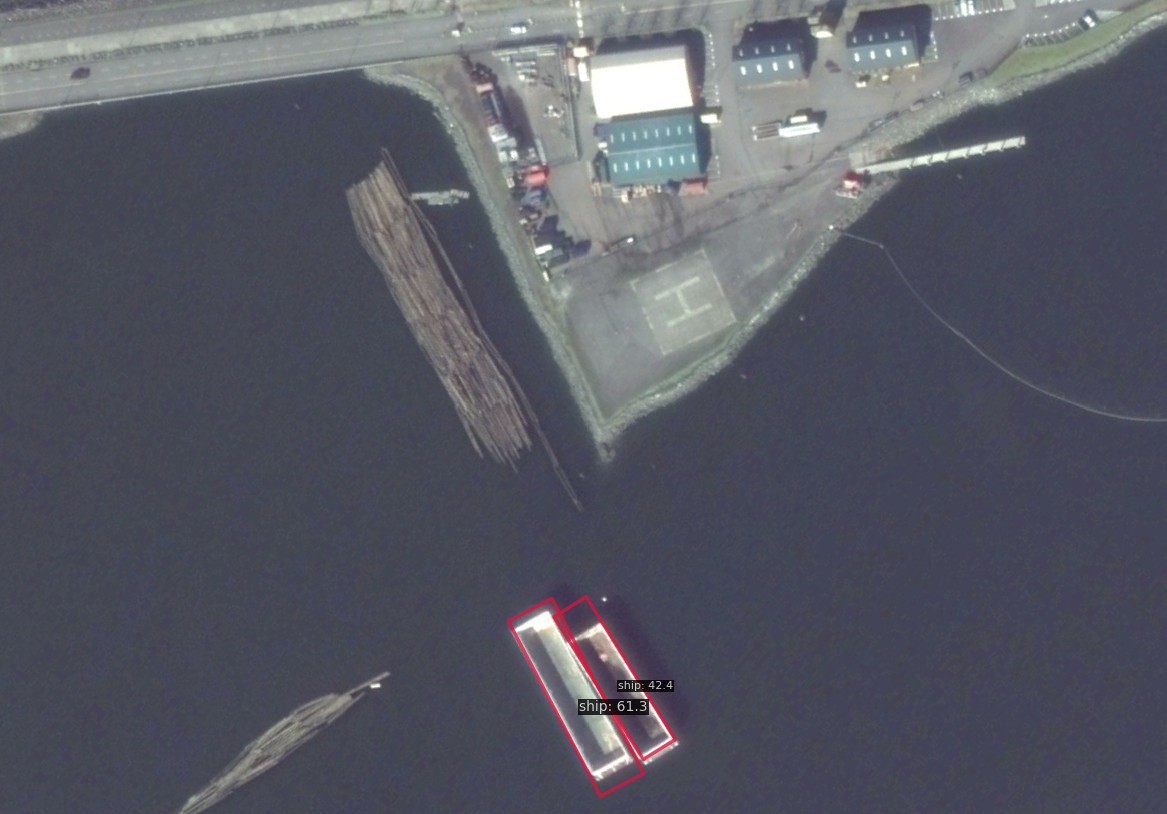}
  \includegraphics[width=0.25\linewidth,height=0.18\linewidth]{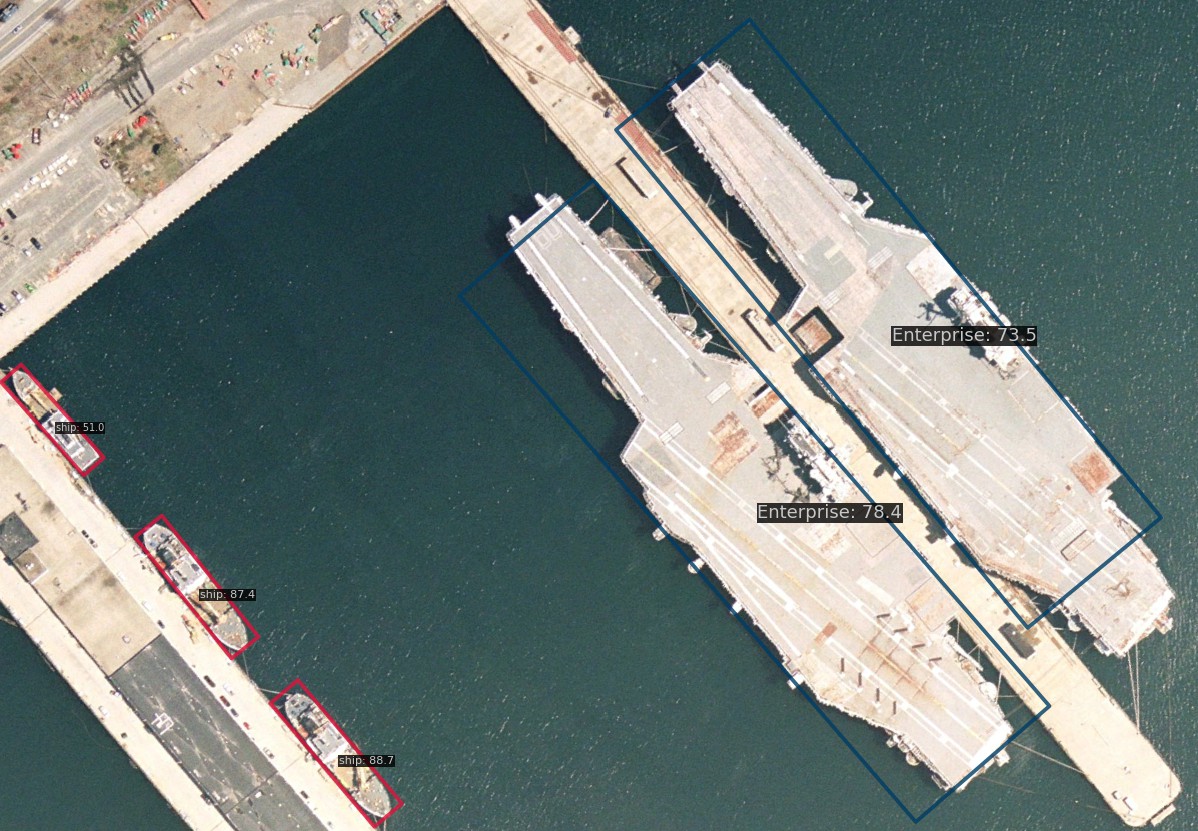}
  }
  \caption{Rotated object detection results from the test set of DOTA \cite{DOTA} and HRSC2016 \cite{hrsc} using RTMDET \cite{lyu2022rtmdet} trained using $\mathcal{L}_{RotatedIoU}$ and $\mathcal{L}_{FPDIoU}$ losses.
  }
  \label{result_DOTA_HRSC}
\end{figure*}

\begin{figure*}[h]
  \subfloat[Rotated object detection results training with $\mathcal{L}_{RotatedIoU}$]{\includegraphics[width=0.25\linewidth,height=0.18\linewidth]{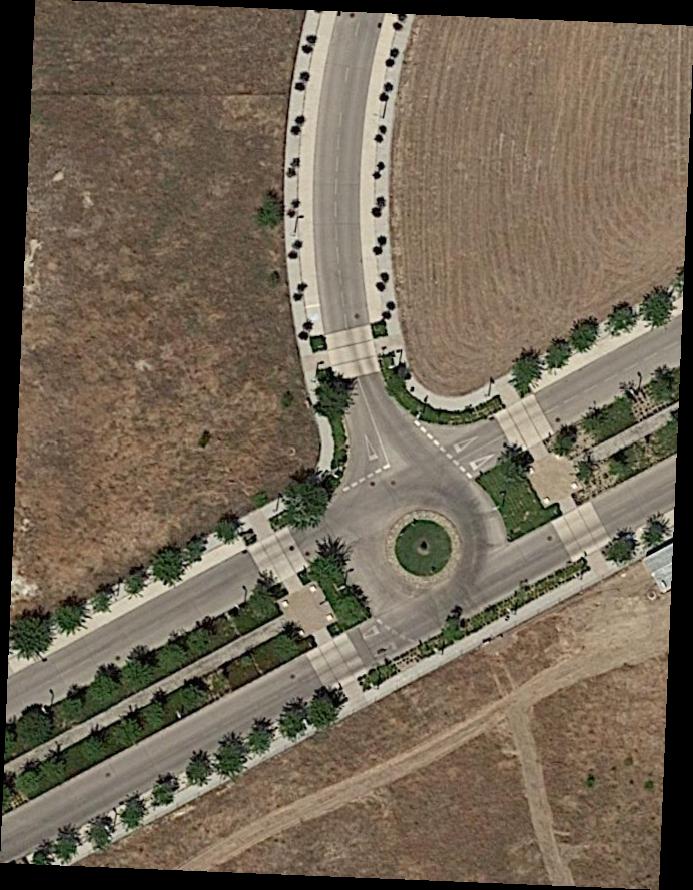}
  \includegraphics[width=0.25\linewidth,height=0.18\linewidth]{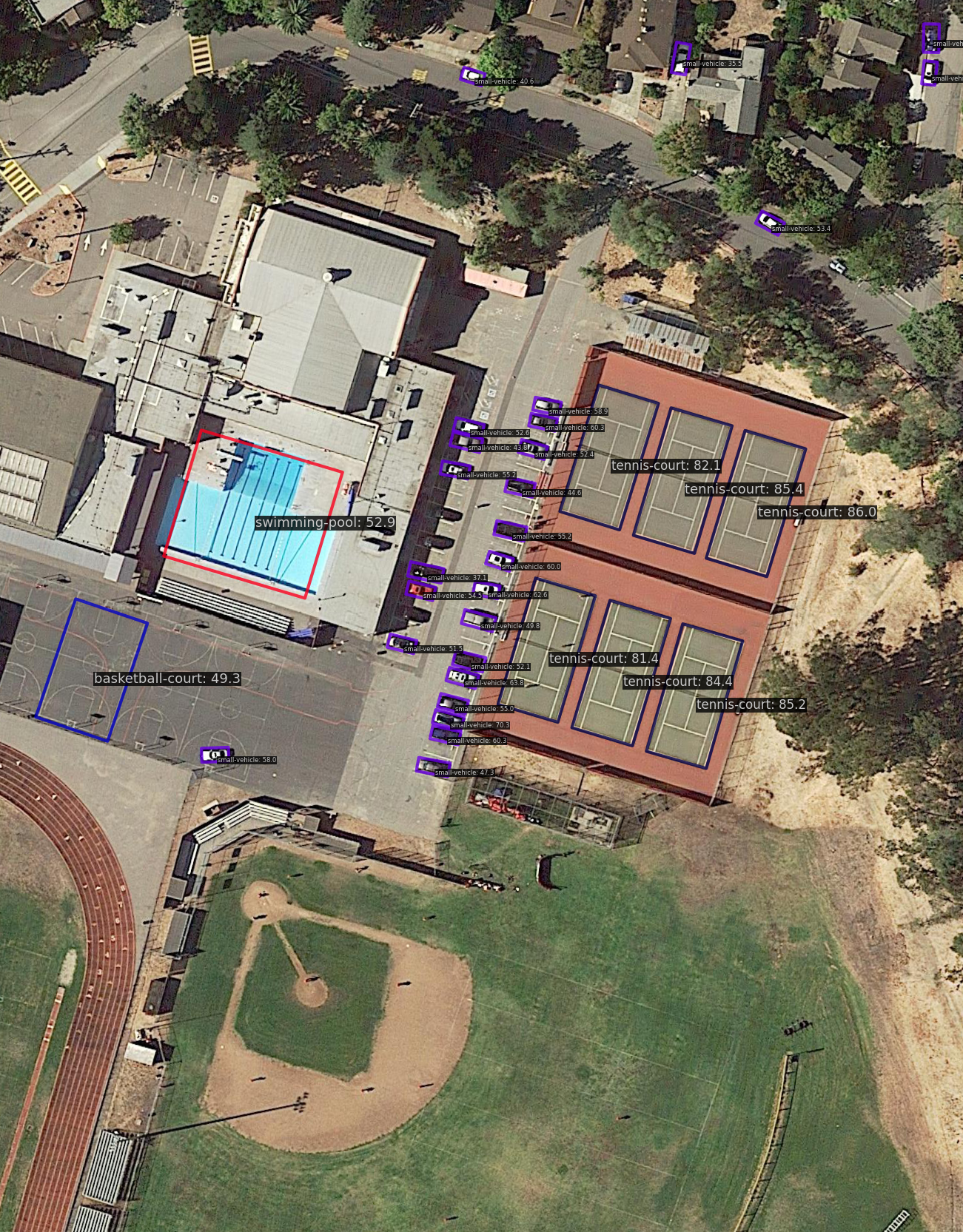}
  \includegraphics[width=0.25\linewidth,height=0.18\linewidth]{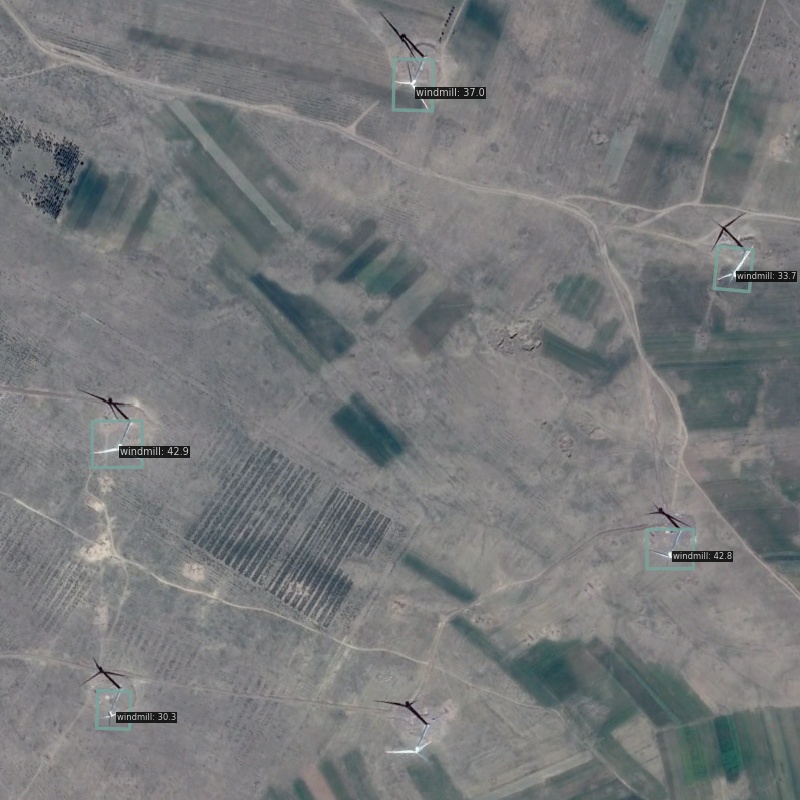}
  \includegraphics[width=0.25\linewidth,height=0.18\linewidth]{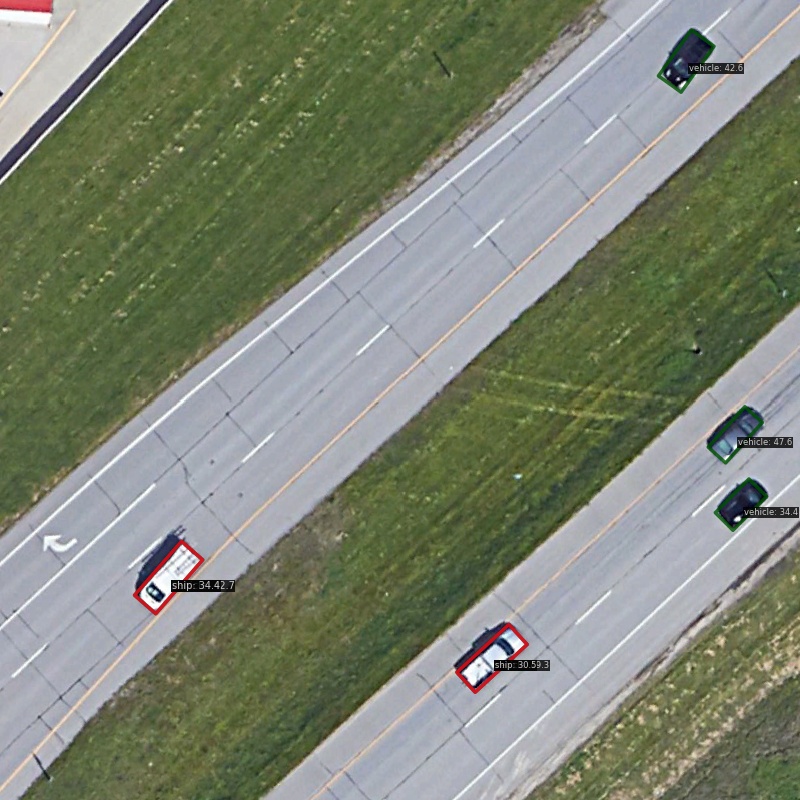}
  
  }\\
  \subfloat[Rotated object detection results training with $\mathcal{L}_{FPDIoU}$]{\includegraphics[width=0.25\linewidth,height=0.18\linewidth]{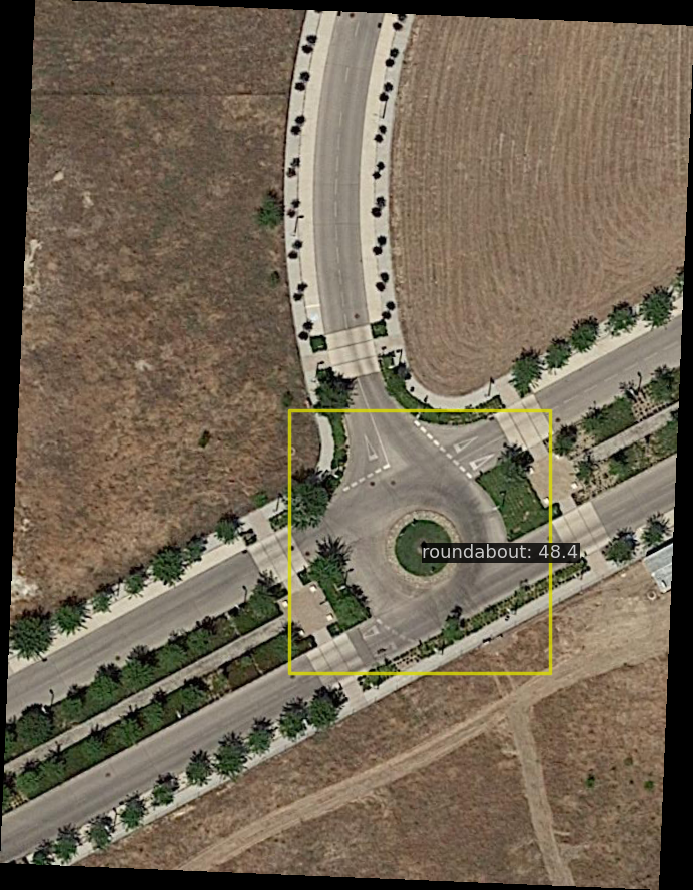}
  \includegraphics[width=0.25\linewidth,height=0.18\linewidth]{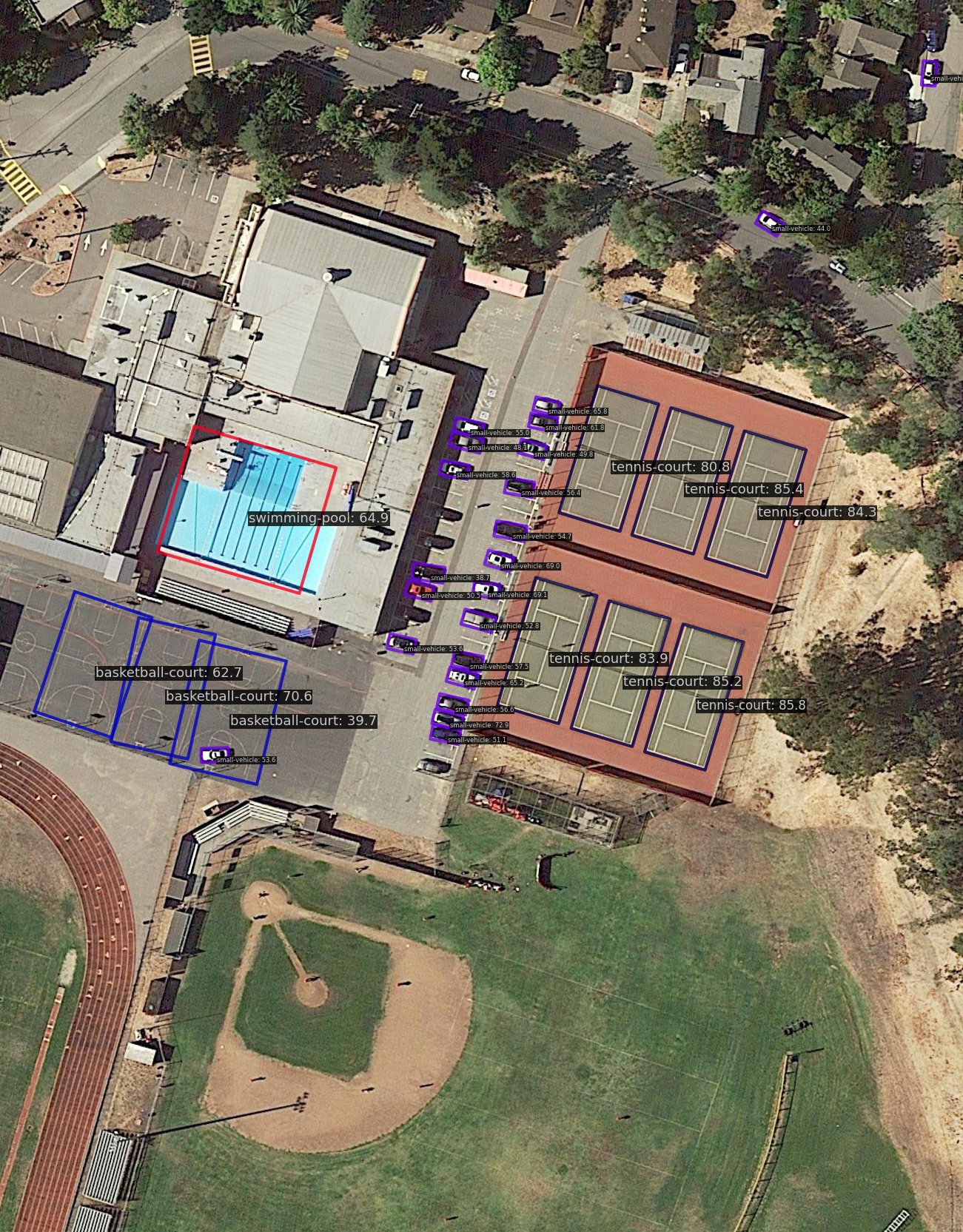}
  \includegraphics[width=0.25\linewidth,height=0.18\linewidth]{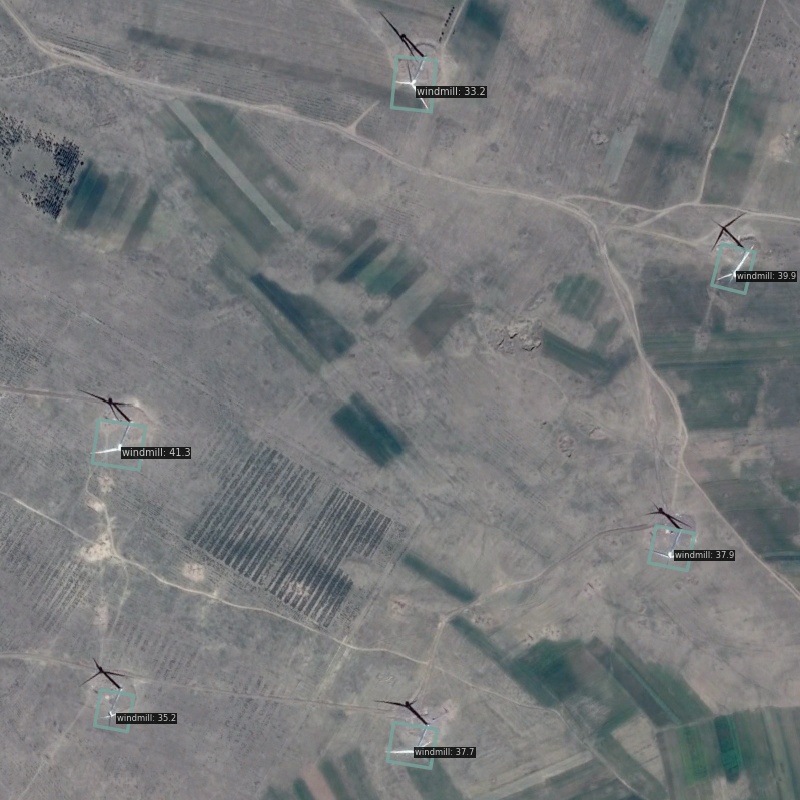}
  \includegraphics[width=0.25\linewidth,height=0.18\linewidth]{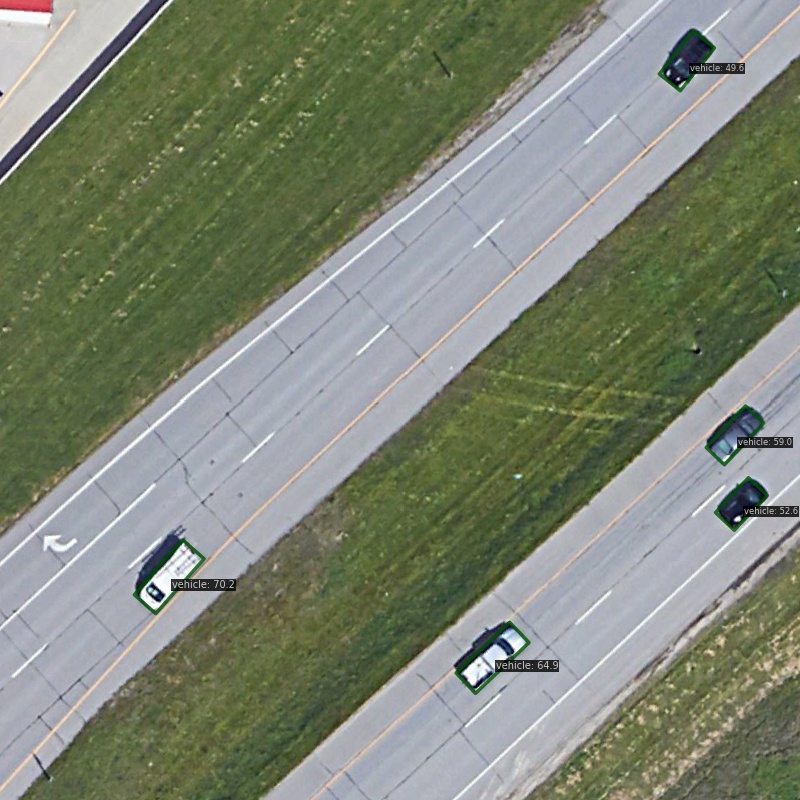}
  }
  \caption{Rotated object detection results from the test set of DOTA \cite{DOTA} and DIOR \cite{DIOR} using H2RBox \cite{yang2022h2rbox} trained using $\mathcal{L}_{RotatedIoU}$ and $\mathcal{L}_{FPDIoU}$ losses.
  }
  \label{result_DOTA_DIOR}
\end{figure*}

\begin{figure*}[b]
  \subfloat[Rotated object detection results training with $\mathcal{L}_{RotatedIoU}$]{\includegraphics[width=0.25\linewidth,height=0.18\linewidth]{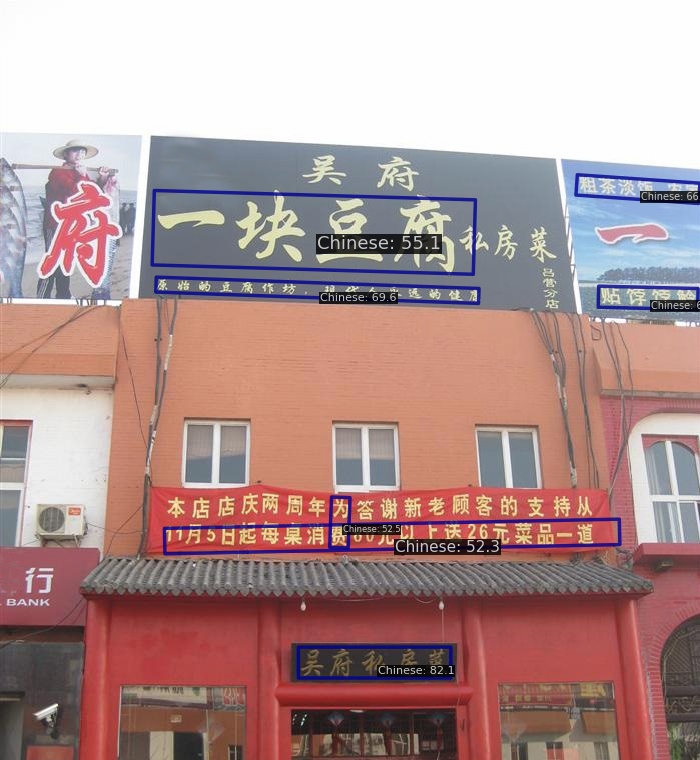}
  \includegraphics[width=0.25\linewidth,height=0.18\linewidth]{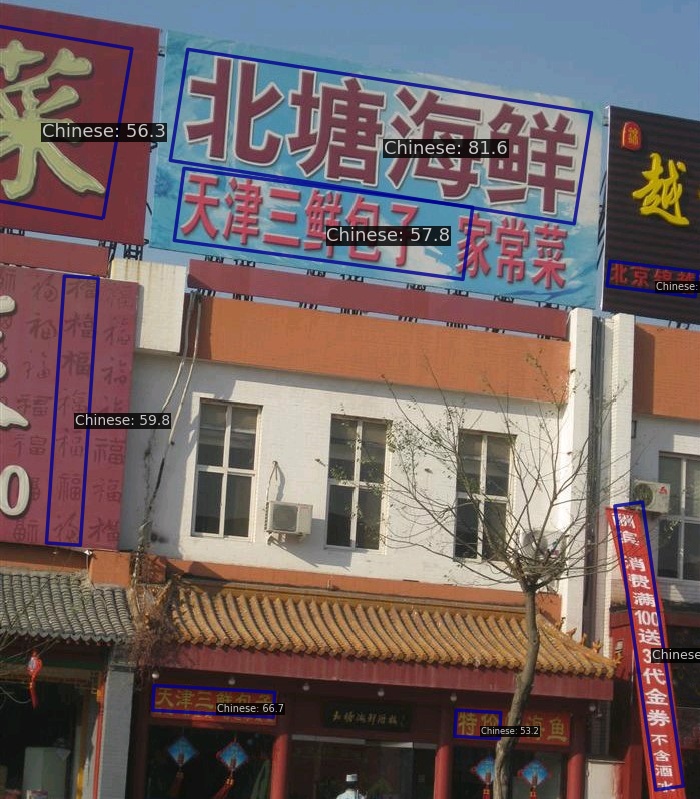}
  \includegraphics[width=0.25\linewidth,height=0.18\linewidth]{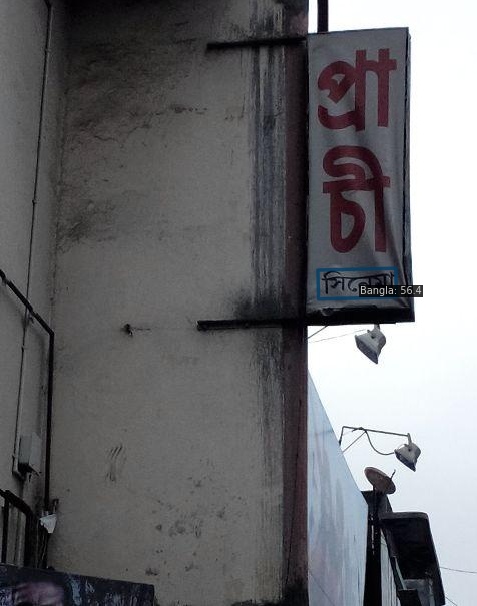}
  \includegraphics[width=0.25\linewidth,height=0.18\linewidth]{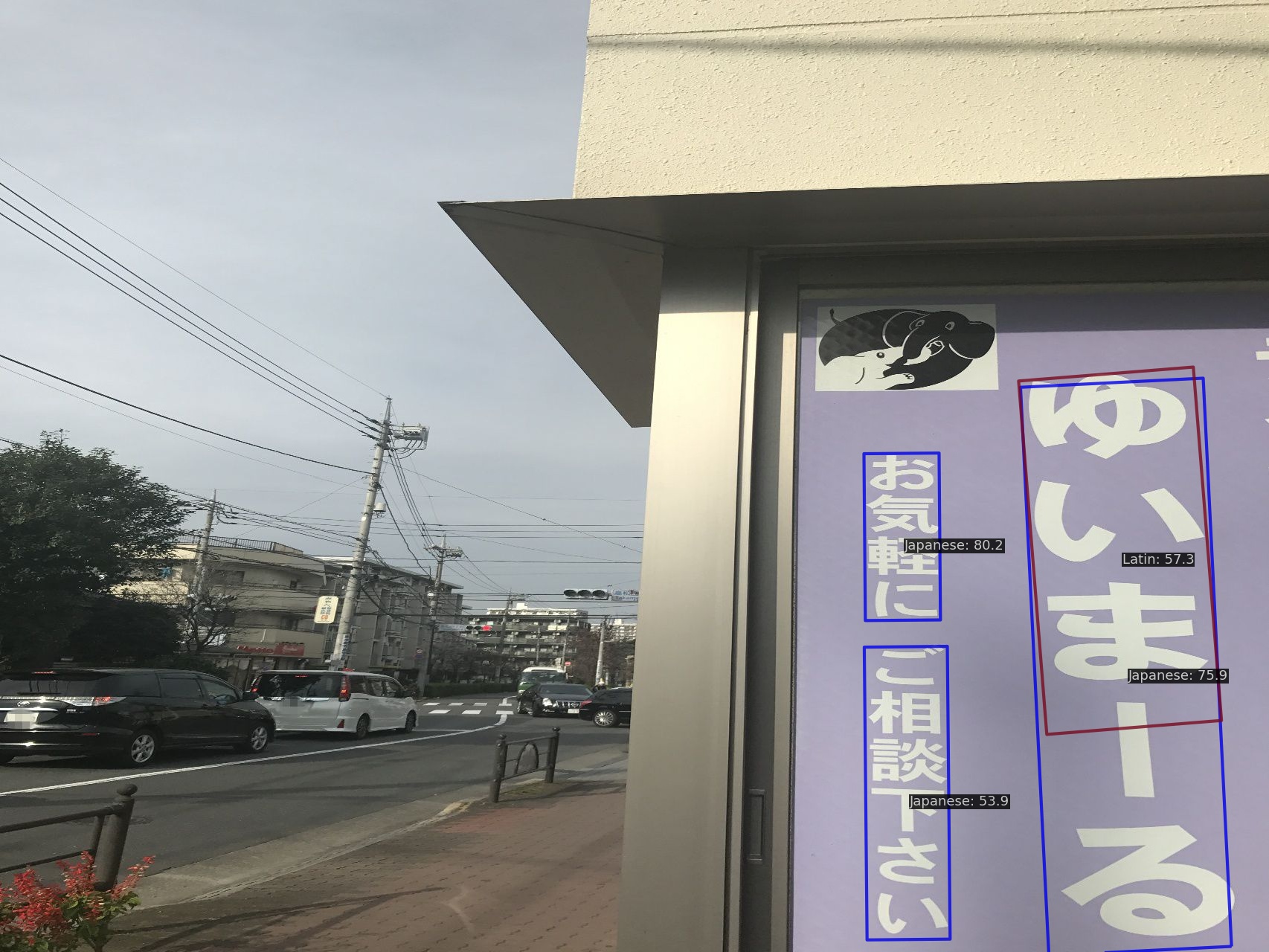}
  
  }\\
  \subfloat[Rotated object detection results training with $\mathcal{L}_{FPDIoU}$]{\includegraphics[width=0.25\linewidth,height=0.18\linewidth]{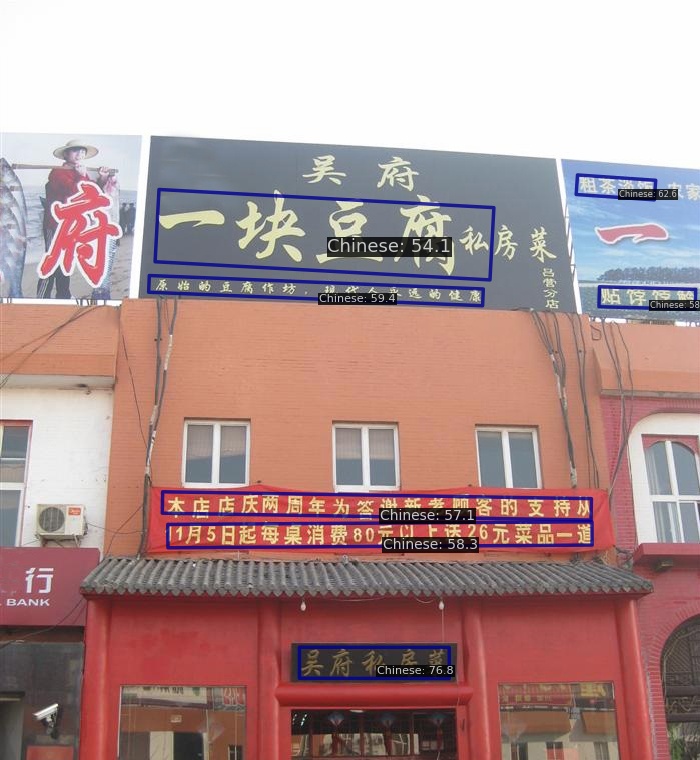}
  \includegraphics[width=0.25\linewidth,height=0.18\linewidth]{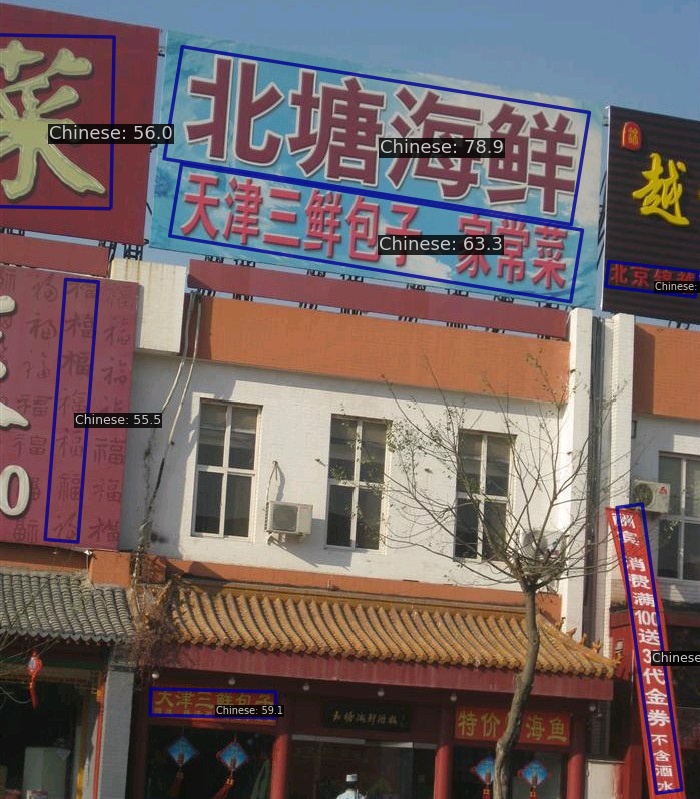}
  \includegraphics[width=0.25\linewidth,height=0.18\linewidth]{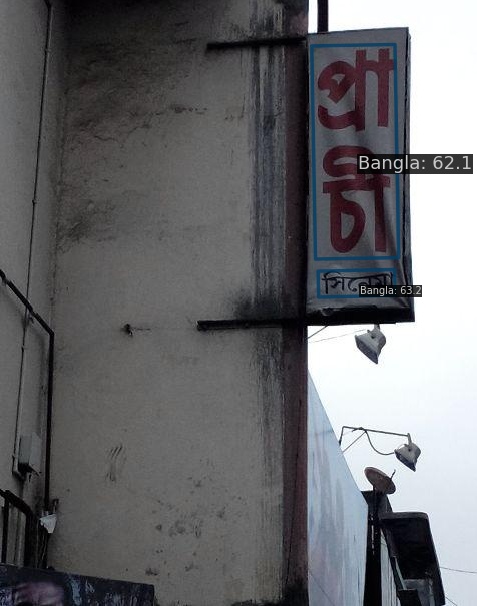}
  \includegraphics[width=0.25\linewidth,height=0.18\linewidth]{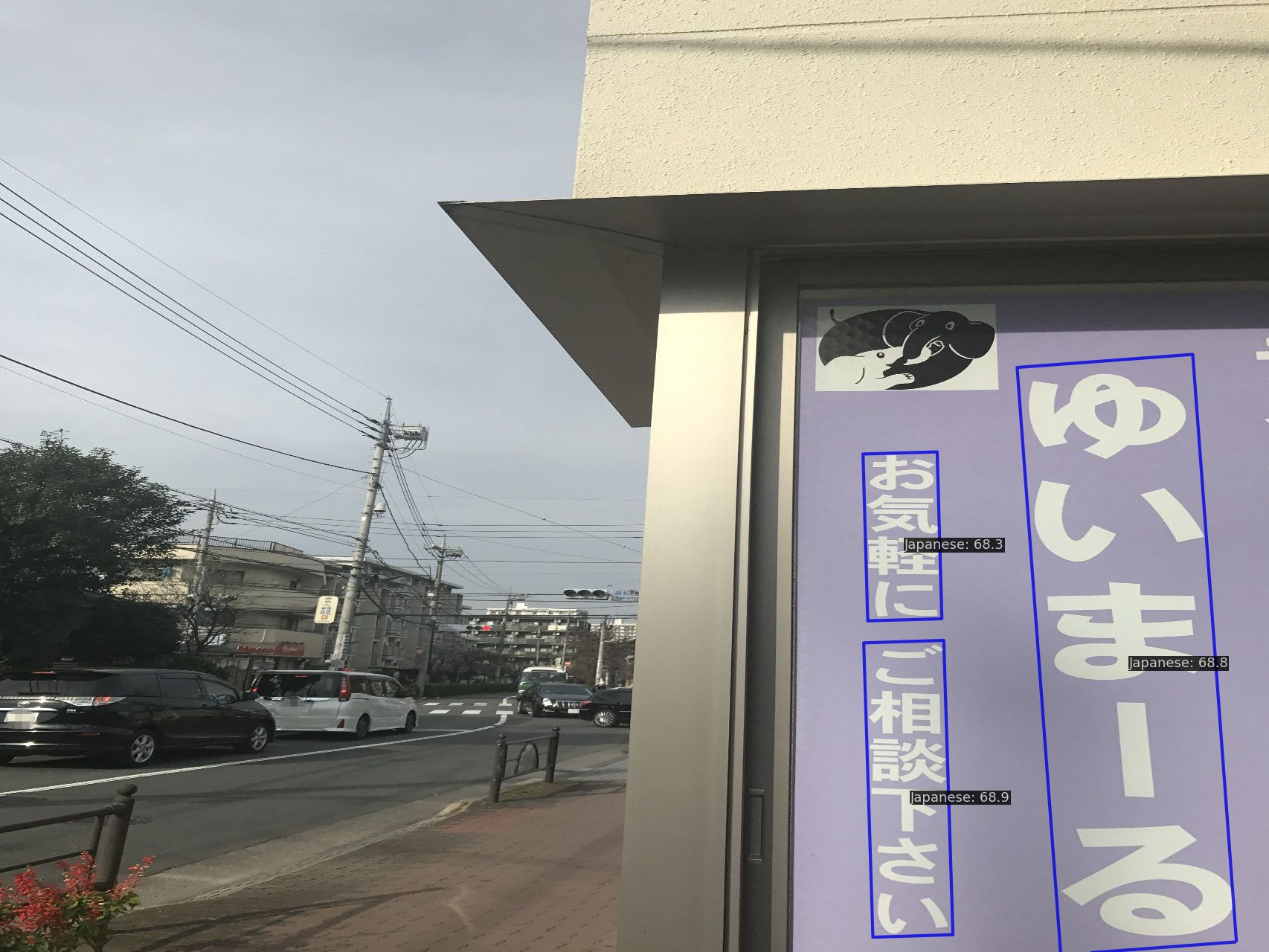}
  }
  \caption{Arbitrary orientation scene text spotting results from the test set of ICDAR 2017 RRC-MLT \cite{nayef2017icdar2017} and ICDAR 2019 RRC-MLT \cite{nayef2019icdar2019} using RTMDET \cite{lyu2022rtmdet} trained using $\mathcal{L}_{RotatedIoU}$ and $\mathcal{L}_{FPDIoU}$ losses.
  }
  \label{result_IC17_IC19}
\end{figure*}

As Figure \ref{result_DOTA_HRSC} and \ref{result_DOTA_DIOR} shows, the rotated object detectors training with our proposed $\mathcal{L}_{FPDIoU}$ achieve better performance than training with their original loss functions. Especially, our visualize results have less redudancy and higher integrity. For example, the terminal was misrecognized as plane in first image of Figure \ref{result_DOTA_HRSC}(a) and the roundabout is not detected in first image of Figure \ref{result_DOTA_DIOR}(a), some trucks were not detected in the second image of Figure \ref{result_DOTA_HRSC}(a). 

In order to better reveal the performance of different loss functions for bounding box regression of rotated object detection, we provide some of the visualization results as Figure \ref{result_DOTA_HRSC} shows. As we can see, the rotated object detection results provided by model training with our proposed $\mathcal{L}_{FPDIoU}$ are with less redudancy and higher accuracy than model training with $\mathcal{L}_{RotatedIoU}$.\\
\textbf{Results of arbitrary orientation scene text spotting}

In order to further verify the effectiveness of the bounding box regression loss function described in this chapter in the arbitrary orientation scene text spotting task, we selected the ICDAR 2017 RRC-MLT and ICDAR 2019 RRC-MLT datasets to be used on the RTMDET \cite{lyu2022rtmdet} model.
The experimental results are shown as Table \ref{rtmdetIC17} and Table \ref{rtmdetIC19}. It can be seen that the performance and efficiency of $\mathcal{L}_{FPDIoU}$ in any direction text detection task proposed in this chapter are still significantly improved, and the accuracy is especially improved, indicating that the $\mathcal{L}_{FPDIoU}$ proposed by ourselves can output more accurate and less redundant text detection results in any direction. As Figure \ref{result_IC17_IC19} shows, the arbitrary orientation scene text spotter training with our proposed $\mathcal{L}_{FPDIoU}$ achieve better performance than training with its original loss functions. Especially, our visualize results have less redudancy and higher integrity. For example, The text in the first red banner in the middle of the first image of Figure \ref{result_IC17_IC19}(a) is not detected and the vertical text is not detected in third image of Figure \ref{result_IC17_IC19}(a). 

\subsubsection{Comparison with the State-of-the-art methods}

In order to verify the performance of our proposed $FPDIoU$, we conduct experiments to compare with the state-of-the-art rotated object detection methods. The experimental results are shown as Table \ref{sota-compare}, which compares recent detectors on DOTA-v1.0, as categorized by single-, refine-, and two-stage based methods. Due to different image resolution, network structure, training strategies and various tricks between different methods, we cannot make absolutely fair comparisons. In terms of overall performance, our method has achieved the best performance so far, at around 81.33\%.

\section{Conclusion}
In this paper, we introduced a new metric named $FPDIoU$ based on minimum points distance for comparing any two arbitrary bounding boxes. We proved that this new metric has all of the appealing properties which existing $IoU$-based metrics have while simplifing its calculation. It will be a better choice in all performance measures in 2D/3D vision tasks relying on the $IoU$ metric.

We also proposed a loss function called $\mathcal{L}_{FPDIoU}$ for bounding box regression. By incorporating it into the state-of-the-art rotated object detection algorithms, we consistently improved their performance on popular rotated object detection benchmarks such as DOTA, DIOR and HRSC2016 using the commonly used performance measures based average precision. Since the optimal loss for a metric is the metric itself, our $FPDIoU$ loss can be used as the optimal bounding box regression loss in all applications which require 2D bounding box regression.

As for future work, we would like to conduct further experiments on some downstream tasks based on rotated object detection, including arbitrary orientation scene text spotting, person re-identification and so on. With the above experiments, we can further verify the generalization ability of our proposed loss functions.

\section*{Declaration of competing interest}
The authors declare that they have no known competing financial interests or personal relationships that could have appeared to influence the work reported in this paper.
\printcredits

\section*{Acknowledgements}
This work was supported by the National Natural Science Foundation of China(No.62072188).

\bibliographystyle{unsrt}

% Loading bibliography database
\bibliography{reference}

% Biography
\parpic[l][r]{\includegraphics[width=1in,height=1.25in,clip,keepaspectratio]{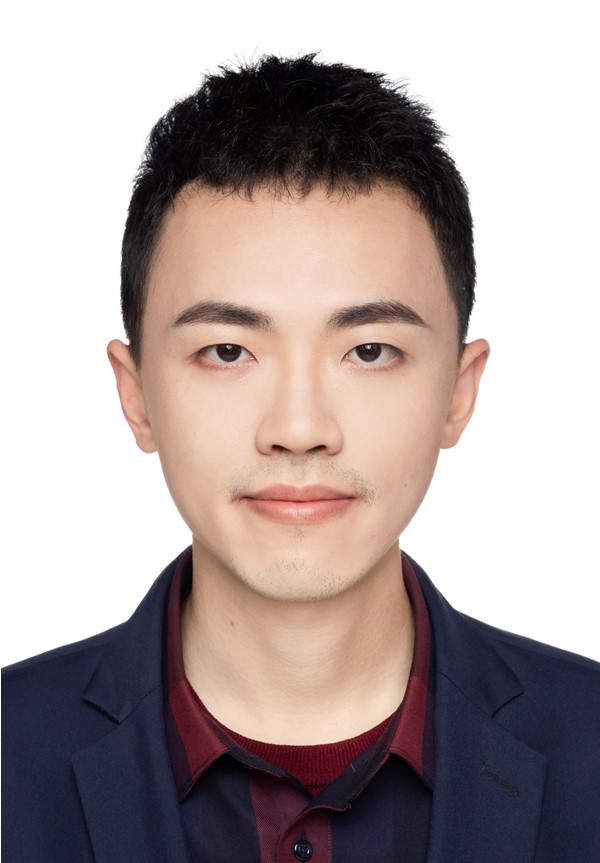}}
\noindent {Siliang Ma is a Ph.D student at the School of Computer Science and Engineering at South China University of Technology. He received his B.E. degree from South China Agricultual University in 2019. He is the President of CCF South China University of Technology Student Chapter. His research interests include deep learning, pattern recognition, and text image processing.}
% \bio{pic1}
% 
% \endbio 

\parpic[l][r]{\includegraphics[width=1in,height=1.25in,clip,keepaspectratio]{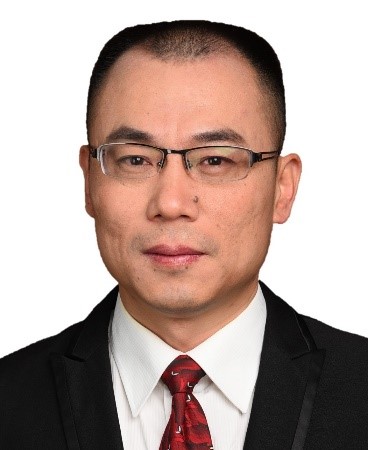}}
\noindent {Yong Xu received the B.S., M.S., and Ph.D. degrees in mathematics from Nanjing University, Nanjing, China, in 1993, 1996, and 1999, respectively. He was a Postdoctoral Research Fellow of computer science with the South China University of Technology, Guangzhou, China, from 1999 to 2001, where he became a Faculty Member and is currently a Professor with the School of Computer Science and Engineering. He is the vice president of South China University of Technology and the Dean of Guangdong Big Data Analysis and Processing Engineering \& Technology Research Center. He is also a member of the Peng Cheng Laboratory. His current research interests include computer vision, pattern recognition, image processing, and big data. He is a senior member of the IEEE Computer Society and the ACM. He has received the New Century Excellent Talent Program of MOE Award.}
% \bio{pic2}
% Yong Xu received the B.S., M.S., and Ph.D. degrees in mathematics from Nanjing University, Nanjing, China, in 1993, 1996, and 1999, respectively. He was a Postdoctoral Research Fellow of computer science with the
% South China University of Technology, Guangzhou, China, from 1999 to 2001, where he became a Faculty Member and is currently a Professor with the School of Computer Science and Engineering. He is the Dean of Guangdong Big Data Analysis and Processing Engineering \& Technology Research Center. He is also a member of the Peng Cheng Laboratory. His current research interests include computer vision, pattern recognition, image processing, and big data. He is a senior member of the IEEE Computer Society and the ACM. He has received the New Century Excellent Talent Program of MOE Award.

\end{document}